\documentclass{article}

     \usepackage[preprint]{neurips_2023}

\usepackage[utf8]{inputenc} 
\usepackage[T1]{fontenc}    
\usepackage{hyperref}       
\usepackage{url}            
\usepackage{booktabs}       
\usepackage{amsfonts}       
\usepackage{nicefrac}       
\usepackage{microtype}      
\usepackage{xcolor}         

\usepackage{microtype}
\usepackage{graphicx}
\usepackage{subfigure}
\usepackage{enumitem}

\usepackage{amsmath}
\usepackage{amssymb}
\usepackage{mathtools}
\usepackage{amsthm}

\title{Evaluation Strategy of Time-series Anomaly Detection with Decay function}

\author{%
  Yongwan Gim\thanks{corresponding author}  
  \And
  Kyushik Min \\
  \AND
  KAKAO corporation\\
  166, Pangyoyeok-ro, Bundang-gu, Seongnam-si, Gyeonggi-do, Republic of Korea \\
  \texttt{\{rowan.gim, queue.min\}@kakaocorp.com}
}

\begin{document}

\maketitle

\begin{abstract}
  Recent algorithms of time-series anomaly detection have been evaluated by applying a Point Adjustment (PA) protocol.
However, 
the PA protocol has a problem of overestimating the performance of the detection algorithms 
because it only depends on the number of detected abnormal segments and their size.
We propose a novel evaluation protocol called the Point-Adjusted protocol with decay function (PAdf) 
to evaluate the time-series anomaly detection algorithm
by reflecting the following ideal requirements:
detect anomalies quickly and accurately without false alarms.
This paper theoretically and experimentally shows that the PAdf protocol solves the over- and under-estimation problems of existing protocols such as PA and PA\%K.
By conducting re-evaluations of SOTA models in benchmark datasets,
we show that 
the PA protocol only focuses on finding many anomalous segments, 
whereas the score of the PAdf protocol considers not only finding many segments but also detecting anomalies quickly without delay.
The codes related to this research are publicly released.
\end{abstract}

\section{Introduction}
\label{intro}

Time-series anomaly detection is the field of detecting unusual patterns or deviations in time-series data. 
The detection algorithms are used to monitor the streaming data in fields where real-time serving is required, such as finance, e-commerce, and advertisement ecosystem (\cite{boniol2021sand, wu2021current, kravchik2018detecting, li2018anomaly, xu2018unsupervised}). 
For example, 
 Yahoo Corporation has released a system that automatically monitors and raises alerts on time series of different properties for different use-cases (\cite{laptev2015generic}).
At Microsoft, time-series anomaly detection has been used to monitor millions of metrics coming from various services (\cite{ren2019time}).
In our company, we employ detection algorithms to monitor various service indicators, including real-time bidding for online advertisements.

Through the experience gained while operating the monitoring system, 
we define that the following three characteristics are required for ideal time-series anomaly detection:
\begin{enumerate}[label=\textbf{R.\arabic*}]
    \item \label{item:1} {\bf The abnormal situations should be detected without missing them.} 
    \item \label{item:2} {\bf The abnormal situations should be detected as soon as possible.}
    \item \label{item:3} {\bf The frequency of false alarms should be low.}
\end{enumerate}

There are two types of anomaly patterns: point anomaly and contiguous anomaly. 
Point anomaly is an anomaly when an outlier value is a single point in a sequence. 
A contiguous anomaly is a set of outlier values as consecutive points in a sequence.
Emergency situations such as a system failure or cyber-attack appear in the form of a contiguous anomaly in time-series data. 
So, early detection of contiguous anomalies is critical in order to terminate failures or attacks early as mentioned in \ref{item:2}.

If anomaly detection occurs in the industrial monitoring system, 
professional operator receives the alert.
Then the alert is reviewed by the operator to identify and respond to the issues.
For this reason, it is more important for detectors to reduce the fatigue of the operator by reducing false alarms than to increase the number of true positives  in contiguous anomaly segments as mentioned in \ref{item:3}.

\begin{figure}[t] \label{fig:toydata}
  \begin{center}
\subfigure[{~Ground truth}]{
\includegraphics[width=0.46\linewidth]{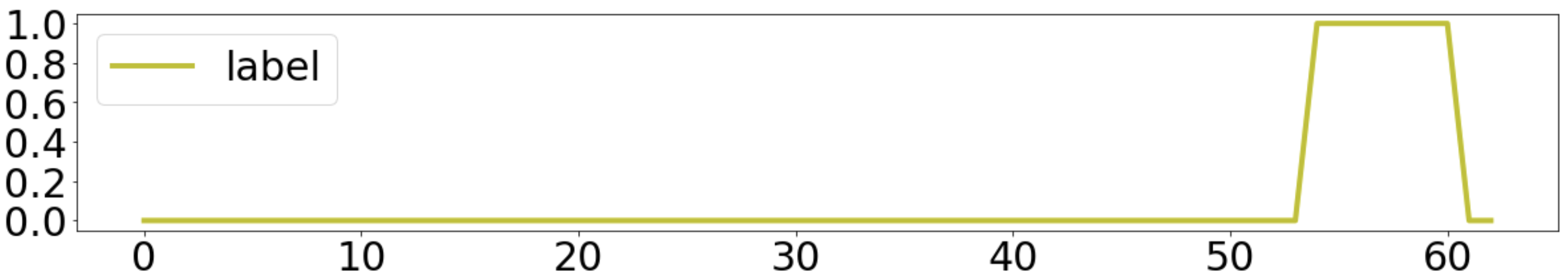}\label{fig:label}} 
\subfigure[{~Anomaly detection case 1}]{
\includegraphics[width=0.46\linewidth]{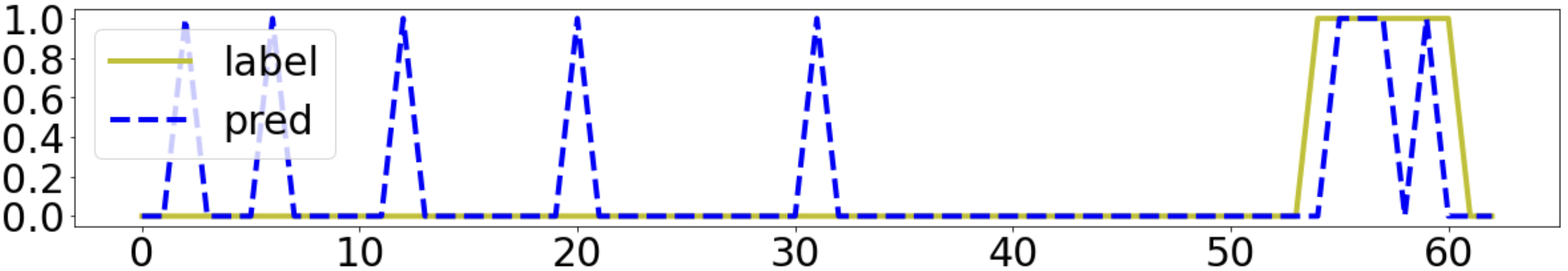}\label{fig:worst}} \\
\subfigure[{~Anomaly detection case 2}]{
\includegraphics[width=0.46\linewidth]{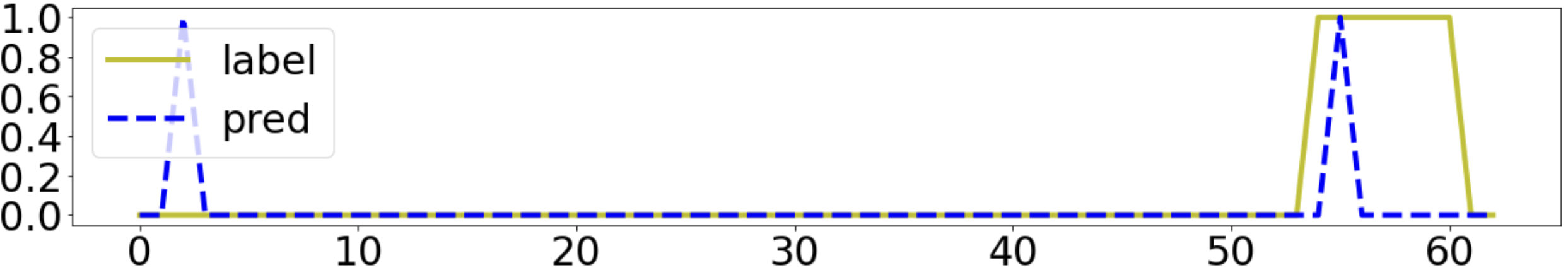}\label{fig:best}} 
\subfigure[{~Anomaly detection case 3}]{
\includegraphics[width=0.46\linewidth]{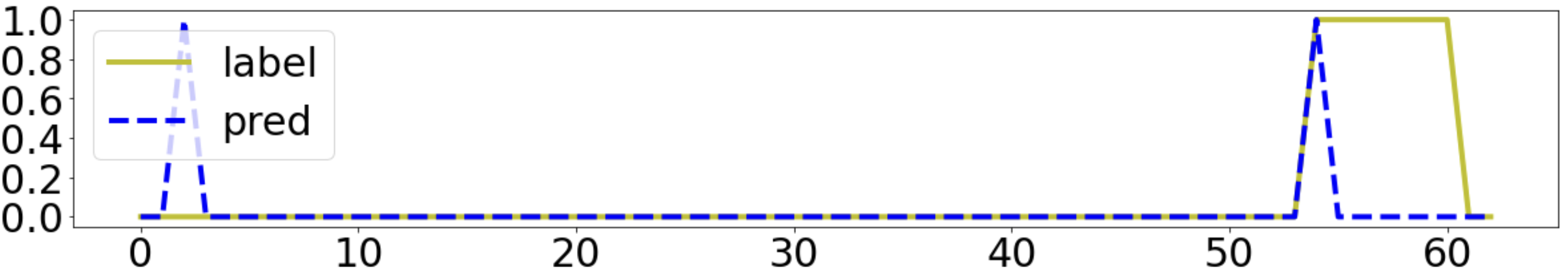}\label{fig:anomaly_case4}} \\
\subfigure[{~Anomaly detection case 4}]{
\includegraphics[width=0.46\linewidth]{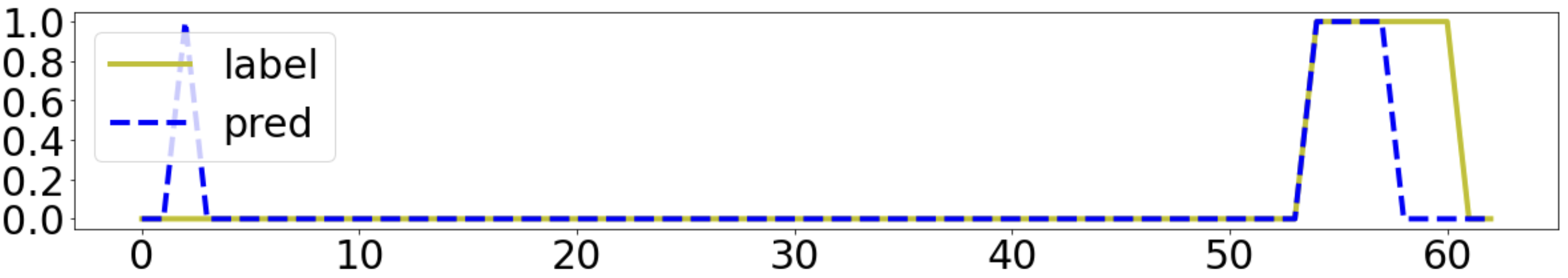}\label{fig:anomaly_case6}} 
\subfigure[{~Anomaly detection case 5}]{
\includegraphics[width=0.46\linewidth]{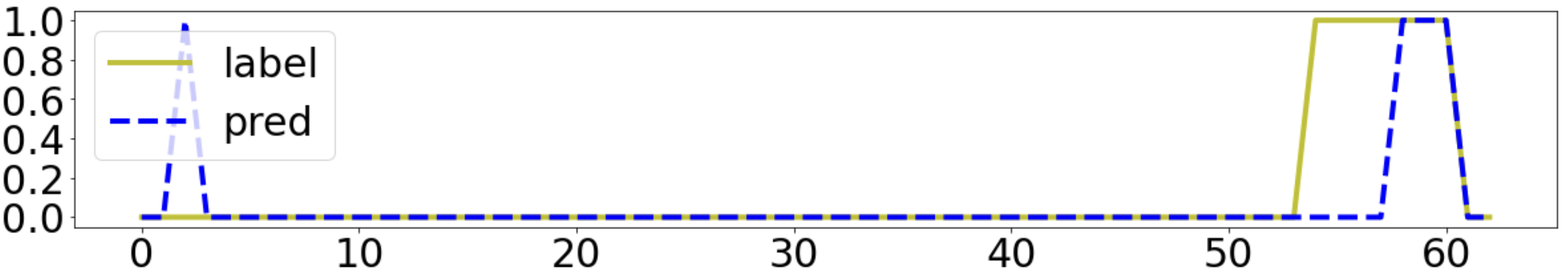}\label{fig:anomaly_case8}} 
  \end{center}
  \caption{Fig.~\ref{fig:label} is the ground truth. Fig.~\ref{fig:worst}$\sim$\ref{fig:anomaly_case8} are the various cases of anomaly detection which can occur in a real-world monitoring system. A detailed explanation is in Sec.~\ref{subsec:toy}.}
\end{figure}


Since time-series detection is a problem of distinguishing anomalies from normal cases, 
the classification evaluation metrics such as precision, recall, and F1 scores can be used to evaluate the anomaly detection algorithms.
However, these metrics give inaccurate evaluation results when the time-series data has a long sequence of the anomaly.
To improve these problems, evaluation methods have been developed in the field of time-series anomaly detection with respect to the various cases, 
such as correction methods for the Numenta benchmark (\cite{lavin2015evaluating}), range precision and recall (\cite{tatbul2018precision}), time-series aware scores (\cite{hwang2019time}), and metric combination (\cite{goswami2022unsupervised}).
Most recently, it has been proposed that the affiliation metric (\cite{huet2022local}) using the duration between the ground truth and the predictions, and the VUS(Volume Under the Surface) (\cite{paparrizos2022volume}), which is an extension of traditional AUC-ROC and AUC-PR measures as a parameter-free and threshold-independent evaluation metric.

\begin{figure}[t]
  \begin{center}
\subfigure[{~Point adjustment}]{
\includegraphics[width=0.46\textwidth]{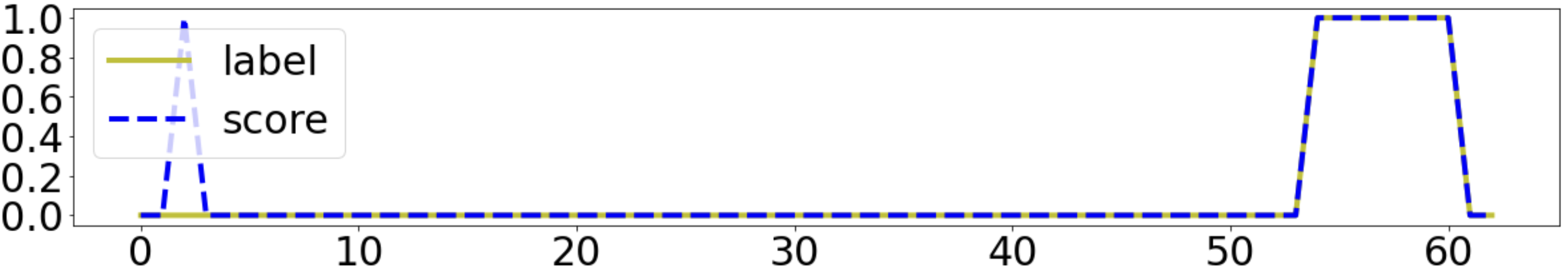}\label{fig:pa1}} \,\,
\subfigure[{~Point adjustment \% K}]{
\includegraphics[width=0.46\textwidth]{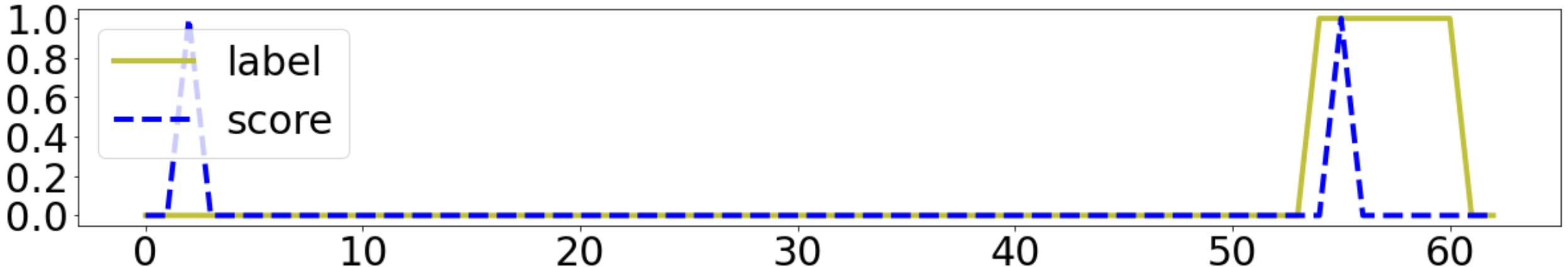}\label{fig:pak}}  \\
\subfigure[{~Point adjustment with decay function}]{
\includegraphics[width=0.46\textwidth]{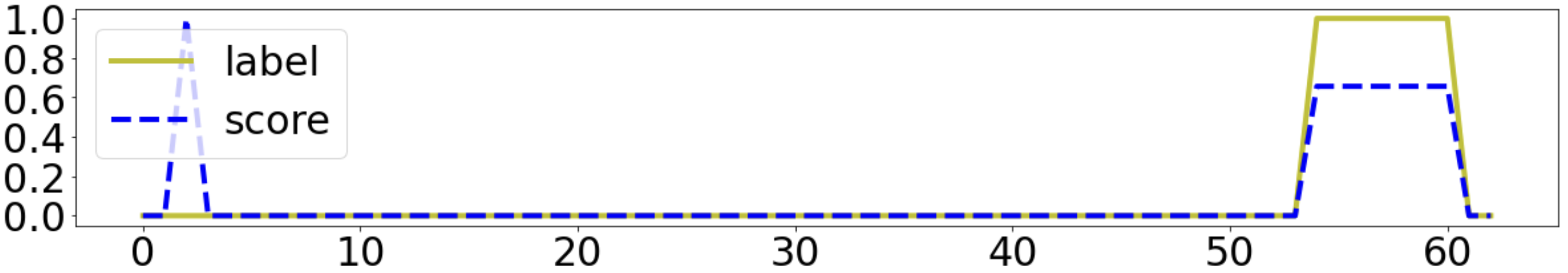}\label{fig:PAdf1}}
  \end{center}
  \caption{The figures showing how the protocols adjust the prediction result when the ground truth is Fig.~\ref{fig:label} and the prediction result is Fig.~\ref{fig:best}, respectively.}\label{fig:case1}
\end{figure}

On the other hand, recent research of the time-series anomaly detection has employed the Point-Adjusted (PA) protocol for model evaluations (\cite{xu2018unsupervised, su2019robust, shen2020timeseries, audibert2020usad, xu2022anomaly}). 
In the PA protocol, if more than one detection occurs in a contiguous anomaly segment, 
 all detection results in the segment are adjusted as a successful detection in order to calculate precision, recall, and F1-score (\cite{xu2018unsupervised}). 
For example, even if an anomaly is detected only once within the contiguous anomaly segment as Fig.~\ref{fig:best}(ground truth: Fig.~\ref{fig:label}), the scores are calculated by adjusting the predicted value as Fig.~\ref{fig:pa1}. 
However, over-estimation phenomena sometimes occur in the PA protocol (\cite{hwang2022you, huet2022local, paparrizos2022volume}).
The over-estimation phenomena mean that 
the PA protocol gives rise to evaluating the performance of the detection model excessively higher than the actual performance.
For example, the performance of random scoring from a uniform distribution is evaluated higher than GDN, OmniAnomaly, LSTM-VAE, DAGMM, and USAD (\cite{kim2022towards}).

In order to compensate for the over-estimation phenomena, the PA\%K protocol is proposed in \cite{kim2022towards}, which applies the PA protocol only when an anomaly of K\% or more is detected in the consecutive anomaly segments. 
For example, even if the detector detects the anomalies in Fig.~\ref{fig:label} as Fig.~\ref{fig:best}, the detection result is not adjusted by the PA protocol as shown in Fig.~\ref{fig:pak},
since the detection is not performed at a certain rate or more in the corresponding segment. 
In that paper, the over-estimation phenomena of PA are shown by mathematical demonstrations, but the fact that PA\%K resolved the overestimation is just experimentally verified.
However, there is an under-estimation problem in the PA\%K protocol. 
For example, even when the contiguous anomaly is detected as soon as it occurs as Fig.~\ref{fig:anomaly_case4}, the F1-score of PA\%K is only 0.25 while F1-score of PA is 1.0 as shown in Table~\ref{table:PAvsPAKvsPADF}. 
But, in the real monitoring system, even though only one anomaly is detected in the anomaly segment, 
the operator receives an alert and responds to the first detection even in the case of Fig.~\ref{fig:anomaly_case4}. 
Therefore, the detection of the anomaly segment should be evaluated as a successful detection because it minimizes the loss of service and business due to anomaly.

On the other hand, both PA and PA\%K (with $K=20$) protocols evaluate F1 score as 1.0
in the case of Fig.~\ref{fig:anomaly_case8} where the anomaly event is detected much later it happened even though the detector detects more than 20\% of anomalies in the anomaly segment. 
However, there are service and business losses caused by delayed detection in continuous anomaly situations, so this is another example of the over-estimation phenomenon.
The reason for this phenomenon is that the PA and PA\%K protocols do not reflect the order of the true positive in the anomaly segment,
which is the key point of \ref{item:2}.

In this paper, 
we propose the Point-Adjusted protocol with decay function (PAdf) 
which is a novel evaluation protocol to evaluate the algorithms for time-series anomaly detection 
by reflecting the ideal requirements \ref{item:1}, \ref{item:2} and \ref{item:3}.

The contributions of PAdf are summarized as follows:
\begin{itemize}
\item We mathematically formulate the PAdf protocol that satisfy \ref{item:1}, \ref{item:2} and \ref{item:3}, and derive the novel form of the precision, recall, F1 score.
\item We show that the PAdf protocol compensates for the problem of over- and under-estimating anomaly detectors in various anomaly situations.
\item By using the PAdf protocol to calculate metrics, the F1 score of recent SOTA models for the time-series anomaly detection is re-evaluated and compared with the results of the metrics from the conventional protocols.
\end{itemize}

\section{Point Adjustment Protocols}
\label{sec:PA}

In this section,
we derive precision and recall from the mathematical definition of the PA protocol.
Then, it shows that the over-estimation phenomenon is able to occur when a random score model is assumed.
Additionally, the PA\%K protocol is also explained.

 Precision $\mathbb{P}$, recall $\mathbb{R}$, F score $\mathbb{F}_\beta$ as the evaluation metrics in the field of the classification are defined as
\begin{align}
\mathbb{P} = \frac{\rm TP}{\rm TP+FP},~ \mathbb{R} = \frac{\rm TP}{\rm TP+FN},~{\mathbb{F}_{\beta}} = (1+\beta^2) \cdot \frac{ {\rm Precision} \cdot {\rm Recall}}{\beta^2 \cdot {\rm Precision} + {\rm Recall}} \label{eq:scores}
\end{align}
where TP, FP, FN, and FP denote the true-positive, false-positive, false-negative, and false-positive, respectively.
The $\beta$ of the ${\mathbb{F}_{\beta}}$ is fixed as $\beta = 1$ in this paper.

The recall $\mathbb{R}$ and precision $\mathbb{P}$ in Eq.~\eqref{eq:scores} are rewritten in terms of the conditional probability as
\begin{align}
\mathbb{R} &= {\cal P}(\hat{y}=1 | y=1) = 1 - {\cal P}(\hat{y}=0 | y=1), \label{eq:R2} \\
\mathbb{P} &= {\cal P}(y=1 | \hat{y}=1) 
= \frac{\mathbb{R}\cdot {\cal P}(y=1)}{\mathbb{R}\cdot {\cal P}(y=1) + {\cal P}(\hat{y}=1, y=0)}, \label{eq:P2}
\end{align}
where ${\cal P}(y)$ is the probability of anomaly at $y$.

\subsection{PA}
\label{subsec:PA_over}

We recall the PA protocol that, if more than one detection occurs in a contiguous anomaly segment, 
all detection in the anomaly segment are adjusted as a successful detection.
The sum of the conditional probability given an anomaly segment $\cal A$ is rewritten 
in terms of the recall $\mathbb{R}$ of PA protocol as 
\begin{align}
1  
= &{\cal P}\left(\forall t \in {\cal A}, S(t) < \theta | t \in {\cal A}\right)   + \mathbb{R}_{\rm PA}, \label{eq:R_PA1}
\end{align}
where $S(t)$ is the anomaly score of the score function from algorithms of anomaly,
and the case where the score is greater than a threshold $\theta$ is considered an anomaly situation by the algorithms.

Then, the recall of the PA protocol can be simply written from Eq.~\eqref{eq:R_PA1} as
\begin{align}
\mathbb{R}_{\rm PA} 
=  1 -  \frac{{\cal P}\left(\forall t \in {\cal A}, S(t) < \theta\right) {\cal P}\left(t \in {\cal A}\right)}{{\cal P}\left(t \in {\cal A}\right)} 
=  1 -  {\cal P}\left(\forall t \in {\cal A}, S(t) < \theta\right)  \label{eq:R_PA_prob},
\end{align}
where we use that
${\cal P}\left(\forall t \in {\cal A}, S(t) < \theta, t \in {\cal A}\right) = {\cal P}\left(\forall t \in {\cal A}, S(t) < \theta\right) {\cal P}\left(t \in {\cal A}\right)$.

\begin{figure}[h]
  \begin{center}
\subfigure[{~Point adjustment protocol}]{
\includegraphics[width=0.46\textwidth]{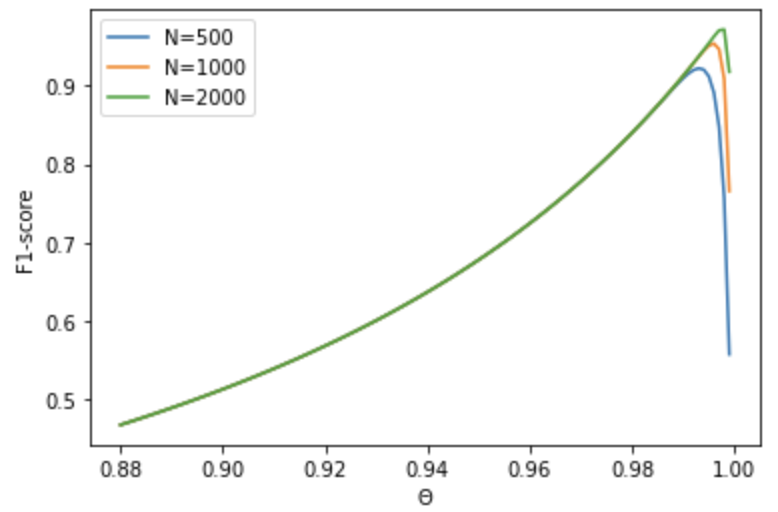}\label{fig:pa}} \,\,
\subfigure[{~Point adjustment with decay function}]{
\includegraphics[width=0.46\textwidth]{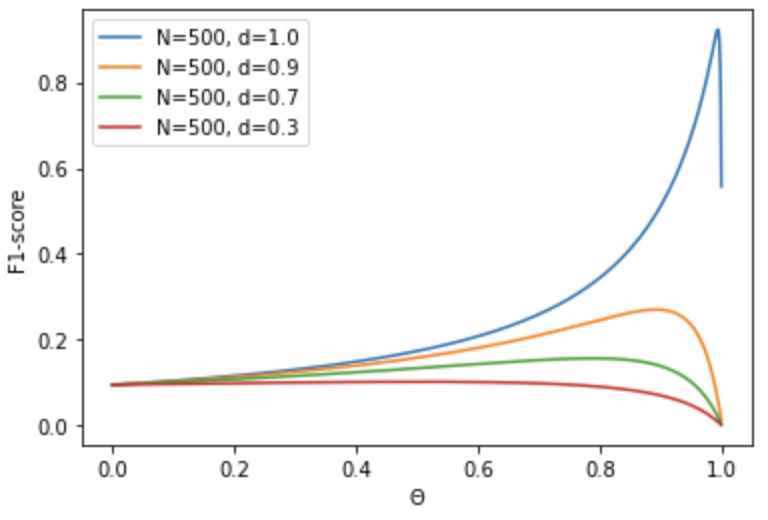}\label{fig:PAdf}}
  \end{center}
  \caption{$N$ means the length of anomaly segment as $N=t_e-t_s,~A = \{ts,t_{s+1}, \cdots,te\}$. The anomaly ratio ${\cal P}_{\cal A}$ is fixed as 0.05. In Fig.~\ref{fig:PAdf}, the case of $N=500,d=1.0$ is the same case as $N=500$ in Fig.~\ref{fig:pa}.}\label{fig:case1}
\end{figure}


Now, we assume the random score model (see Sec.~\ref{subsub:baseline_bench}) in order to show the over-estimation phenomenon.
Then, the probability of a normal situation is given as ${\cal P}(S(t) < \theta) = \theta$,
so that the recall is obtained as
\begin{align}
\mathbb{R}_{\rm PA} 
= 1 - \theta^N.  \label{eq:SKimpaper} 
\end{align}
It is worth noting that the definition of the protocol is mathematically reflected in Eq.~\eqref{eq:SKimpaper}:
all timestamps in the anomaly segment $\cal A$ are considered correctly detected except for the case where all timestamps in the segment predict as a normal situation.

In addition, 
the precision is rewritten in terms of the recall by the use of Eq.~\eqref{eq:P2} as
\begin{align}
\mathbb{P}_{PA}= \frac{\mathbb{R}_{\rm PA} \cdot {\cal P}_{\cal A}}{\mathbb{R}_{\rm PA} \cdot {\cal P}_{\cal A} + (1-\theta)(1-{\cal P}_{\cal A})}
\end{align}
where ${\cal P}_{\cal A}$ means a proportion of anomalies in total time-series data as ${\cal P}_{\cal A} = {\cal P}(t \in {\cal A})$.
Using Eq.~\eqref{eq:scores}, the F1 score with respect to the threshold $\theta$ is shown in Fig.~\ref{fig:pa} which corresponds to the graph in \cite{kim2022towards}.

\subsection{PA\%K}

PA\%K is the most recently developed evaluation method that complements the limitations of PA (\cite{kim2022towards}).
Specifically, PA\%K is experimentally shown that the over-estimation phenomenon of the F1 score of the PA protocol is mitigated.
The key point of the PA\%K protocol is that the PA protocol is applied only when the ratio of the number of TP within an anomaly segment exceeds a threshold $K$.
We use this protocol as a baseline for the evaluation metric of the anomaly detection models.
To get the experimental result, we use the GitHub code implementation provided in \cite{kim2022towards}.

\section{Point Adjustment with Decay Function}
\label{sec:PAdf}

Now, let us formulate the mathematical form of the PAdf protocol.
The key to PAdf is that it attenuates TP when the anomaly segments are discovered late.
After detecting the anomaly segment at $t=t_0$, 
TP is scored as having detected all of it with a decayed score as ${\rm TP} = D(t_0)N$ with the length of the anomaly segment $N$,
as shown in Fig.~\ref{fig:PAdf1}.
So, since the sequence of the anomalies in the segment is important in the PAdf protocol, we start with the definitions as
\begin{itemize}
  \item {\bf Set of the anomaly segment} ${\cal A} = \{t_s < t_{s+1}< t_{s+2}< \cdots < t_e\}$ with $|{\cal A}| = N$\\
  \item {\bf Decay function} $D(t)$ is a decreasing function as $t$ goes to $t_e$ from $t_s$ with $D(t_s)=1$.
\end{itemize}

The adjusted probability $\Tilde{{\cal P}}$ can be defined 
by reflecting the decay function 
on the probability of finding anomalies at each point in time.
First, the probability of not detecting an anomaly segment at all is defined as 
\begin{align}
\Tilde{\cal P}&\left(\forall t \in {\cal A}, S(t) < \theta , t \in {\cal A}\right) =  \prod^N_{i=0} p(t_i)  {\cal P}_{\cal A}, \notag
\end{align}
where $p(t)$ is the probability that the detection algorithm predicts the data at time $t$ as normal, that is $p(t) \equiv {\cal P}(S(t) < \theta)$ with the threshold $\theta$.
If the detection algorithm gives an output as a probability, then $p(t)$ is able to be considered as the output of the detection algorithm.

Next, when the algorithm detects the anomaly segment at time $t_{s}$, the probability is rewritten in terms of the decay function as,
\begin{align}
\Tilde{{\cal P}}&\left(S(t_s) > \theta, t \in {\cal A}\right)={\cal D}(t_s)  \left(1- p(t_s)\right)  {\cal P}_{\cal A}, \notag 
\end{align}
where the adjusted probability is the same as that of the PA protocol, since ${\cal D}(t_s) = 1$.
However, if the anomaly in the segment is detected at the time $t_{s+1}$, the adjusted probability is given as follows
\begin{align}
\Tilde{\cal P}\left(S(t_s) < \theta, S(t_{s+1}) > \theta, t \in {\cal A}\right) 
= {\cal D}(t_{s+1}) p(t_s)\left(1- p(t_{s+1})\right) {\cal P}_{\cal A}. \notag 
\end{align}
Then, the effective true positives (eTP) are obtained by the sum of the adjusted probabilities as 
\begin{align}
{\rm eTP} 
&= \tau {\cal P}_A \sum^{N-1}_{n=0}{\cal D}(t_{s+n}) (1-p(t_{s+n})) \prod^{n-1}_{i=0} p(t_{s+i}) 
\end{align}
with the total input size $\tau$,
where the product $\prod^m_n$ with $m<n$ is the empty product as $\prod^m_n (\cdots)=1$.

Finally, the recall of the PAdf protocol is derived as
\begin{align}
\mathbb{R}_{\rm PAdf} = \sum^{N-1}_{n=0}{\cal D}(t_{s+n}) (1-p(t_{s+n})) \prod^{n-1}_{i=0} p(t_{s+i}),  \label{eq:RPAdf1}
\end{align}
and the precision in terms of decay function is also obtained as
\begin{align}
\mathbb{P}_{\rm PAdf} = \frac{{\cal P}_{\cal A} \cdot \mathbb{R}_{\rm PAdf}}{\tau^{-1}\sum^{\tau-N}_{t\in {\cal A}^{\rm c}}(1-p(t))+\left(1-\prod^N_{i=0}p(t_i)\right){\cal P}_{\cal A}}, \label{eq:PPAdf1}
\end{align}
where we use that
\begin{align}
{\cal P}(\hat{y}=1, y=0) = \tau^{-1}\sum^{\tau-N}_{t\in {\cal A}^{\rm c}}(1-p(t)),\quad
{\cal P}(\hat{y}=1, y=1) = \left(1-\prod^N_{i=0}p(t_i)\right){\cal P}_{\cal A}.\notag
\end{align}

Additionally, in the case that the decay function is constant as ${\cal D}(t) =1$, Eq.~\eqref{eq:RPAdf1} becomes the recall of point adjustment protocol in Eq.~\eqref{eq:R_PA_prob} as
\begin{align}
\mathbb{R}_{\rm PAdf} 
= \sum^{N-1}_{n=0} (1-p(t_{s+n})) \prod^{n-1}_{i=0} p(t_{s+i}) 
= 1 - \prod^{N-1}_{i=0}p(t_i), \label{eq:R_PA_reprod}
\end{align}
where $t_{N-1} = t_e$.

From now on, the exponential decay function is defined as
\begin{align} 
    {\cal D}(t) = d^{t_s - t}, \label{eq:decay_func}
\end{align}
where $t \in {\cal A}$ and the decay rate $d$ is constant with $d \in (0,1]$.

Let us assume that $p(t)$ is the probability from the random scoring model, which is given as $p(t) = \theta$, and $D(t)$ is the exponential decay function in order to analyze the overestimation problem in the case of the PAdf protocol and comparing with the previous results shown in Sec.~\ref{subsec:PA_over}.
The recall of the PAdf protocol is obtained as
\begin{align}
\mathbb{R}_{\rm PAdf} = \sum^{N-1}_{i=0} d^i  (1-\theta) \theta^{i} =\frac{(1-\theta)(1-\theta^i d^i)}{1-\theta d}\label{eq:RPAdf}
\end{align}
and then, the precision is derived as 
\begin{align}
\mathbb{P}_{\rm PAdf} 
= &\frac{{\cal P}_{\cal A} \cdot \mathbb{R}_{\rm PAdf}}{(1-\theta)(1- {\cal P}_{\cal A}) + (1-\theta^N){\cal P}_{\cal A} }. \label{eq:RPAdf}
\end{align}
Here, we note that ${\rm TP}+{\rm FP} = {\rm eTP} + {\rm eFP}$.
In order to compare the behavior of F1 score,  
the F1 score is calculated by the Eq.~\eqref{eq:RPAdf} and Eq.~\eqref{eq:scores} with $\beta=1$,
and plot it as shown in Fig.~\ref{fig:PAdf}.
The larger decay gives rise to the smaller upper bound,
so that
F1 score of the PAdf protocol has an upper bound that is much less than 1.
By comparing Fig.~\ref{fig:pa} and Fig.~\ref{fig:PAdf}, 
it is known that the over-estimation phenomenon does not occur when the PAdf protocol is employed.
Note that we show that the PAdf protocol solves the under-estimation phenomenon of PA\%K in Sec.~\ref{subsec:toy_result}.
Additionally, F1 scores in the case of various probability
distributions are shown in Appendix.~\ref{Appendix:mathVariousPAdf}. 

\section{Experimental Settings}
\label{sec:experiments}

\subsection{Toy data}
\label{subsec:toy}

Fig.~\ref{fig:toydata} shows various cases which can occur in real-world monitoring systems even though they are simple examples.
Fig.~\ref{fig:worst} is a detected case that sends a frequent false positive alarm but detects anomalies many times in the anomaly segment.
On the other hand, Fig.~\ref{fig:best} is a model with low alarm frequency for both false positive and true positive overall, but the first detection for the anomaly segment is reproduced to be the same with Fig.~\ref{fig:worst}. 
For reference, alarm frequency and precision are sometimes in a trade-off relationship depending on the threshold even in the same detection model.

Additionally, we import various toy data with varying anomaly detection timing and frequency within anomaly segments.
Fig.~\ref{fig:anomaly_case4} has the same alarm frequency as Fig.~\ref{fig:best}, but the first detection for the anomaly segment is faster than Fig.~\ref{fig:best}. Fig.~\ref{fig:anomaly_case6} has the same first detection timing as Fig.~\ref{fig:anomaly_case4}, but after that, it detects many anomalies in the segment. 
Fig.~\ref{fig:anomaly_case8} is a case that the anomaly within the segment is performed more than Fig.~\ref{fig:best} and Fig.~\ref{fig:anomaly_case4}, but the first detection is later than them.

\subsection{Benchmark data}
\label{subsec:benchmark}

\subsubsection{Datasets}

\textbf{Secure water treatment (SWaT)} (\cite{goh2017dataset})
 is a dataset to study a modern industrial control system for security.
 The size of training data is 495,000 and that of test data is 449,919 with 51 features. The test data has 12.1\% anomalies.

\textbf{Mars Science Laboratory (MSL)} and \textbf{Soil Moisture Active Passive (SMAP)} (\cite{hundman2018detecting})
datasets are the real-world and expert-labeled data coming out of incident surprise and anomaly reports.
MSL dataset contains 58317 rows of training data and 73729 rows of test data
with 10.5\% anomalies, and the dimension of input features is 57.
SMAP dataset has 135,183 rows of training data and 427,617 rows of test data
with 10.5\% anomalies, and the dimension of input features is 25.
Note that the size of their test data is larger than the size of their training data, respectively.

Additional evaluations of more models with respect to various datasets(HEX/UCR (\cite{UCRArchive2018}) etc) are included in Appendix~\ref{Appendix:moreExp}. 

\begin{table}[t]
\caption{{\bf PAdf} is calculated with two decay rates as $d=0.7, 0.9$ which are denoted as 
${\rm PAdf}_{0.7}$ and ${\rm PAdf}_{0.9}$, respectively. The parameter of PA\%K is fixed as $K=20$. Note that {\bf w/o} denotes the F1 score without any protocols.}
\label{table:PAvsPAKvsPADF}
\vskip 0.15in
\begin{center}
\begin{small}
\begin{sc}
\begin{tabular}{lccccc}
\toprule
Figure & w/o & PA & PA\%K &  ${\rm PAdf}_{0.7}$ &  ${\rm PAdf}_{0.9}$ \\
\midrule
\ref{fig:worst} & 0.500   & 0.736 & 0.736 & 0.580 & 0.689\\
\ref{fig:best}  & 0.222  & 0.933 & 0.222 & 0.760 & 0.881 \\
\ref{fig:anomaly_case4}  &0.222  & 0.933 & 0.222 & {\bf 0.933} & 0.933  \\
\ref{fig:anomaly_case6}  & 0.667 & 0.933 & 0.933 & {\bf 0.933} & 0.933 \\
\ref{fig:anomaly_case8} &0.545  &  0.933 & 0.933 & 0.347 & 0.729 \\
\bottomrule
\end{tabular}
\end{sc}
\end{small}
\end{center}
\vskip -0.1in
\end{table}
\begin{table*}[h]
\caption{{\bf PAdf} is calculated with two decay rates as $d=0.7, 0.9$ which are denoted as 
${\rm PAdf}_{0.7}$ and ${\rm PAdf}_{0.9}$, respectively. 
The parameters $K$ of PA\%K are fixed as $K=20$. Note that {\bf w/o} denotes the F1 score without any protocols.}\label{table:benchmark}
\vskip 0.15in
\begin{center}
\begin{sc}
\begin{tabular}{llccccc}
\toprule
dataset & model & w/o & PA & PA\%K & ${\rm PAdf}_{0.7}$ &  ${\rm PAdf}_{0.9}$ \\
\midrule
MSL 
& Random score & 0.191 & 0.907 & 0.475 & 0.306 &  0.437 \\ %
{} & USAD & 0.229 & 0.760 & 0.474 & 0.442 &  0.473\\ %
{} & GDN & 0.205 & 0.829 & 0.409 & 0.353 &  0.453 \\ %
{} & AnomalyTrans & 0.191 & 0.943 & 0.190 & 0.173  & 0.377\\ %
\hline
SMAP 
& Random score & 0.227 & 0.618 & 0.496 & 0.312 &  0.454 \\ 
{} & USAD & 0.270 & 0.566 & 0.298 & 0.243 &  0.246\\ %
{} & GDN & 0.227 & 0.509 & 0.318 & 0.180 &  0.248 \\ %
{} & AnomalyTrans & 0.227 & 0.882 & 0.410 & 0.261  & 0.352\\ %
\hline
SWaT 
& Random score & 0.217 & 0.453 & 0.453 & 0.371 & 0.412 \\ 
{} & USAD & 0.739 & 0.548 & 0.548 & 0.287 & 0.287\\%
{} & GDN & 0.681 & 0.552 & 0.520 & 0.217 & 0.217\\ %
{} & AnomalyTrans & 0.217 & 0.917 & 0.217 & 0.097 & 0.123\\ %
\bottomrule
\end{tabular}
\end{sc}
\end{center}
\vskip -0.1in
\end{table*}

\subsubsection{Baselines}
\label{subsub:baseline_bench}

Results in Sec.~\ref{sec:results} are conducted by the use of the following models
(detailed in Appendix.~\ref{Appendix:Baselines}).

\textbf{Random score} is randomly generated from  a uniform
distribution such as  $p(t) \sim U(0,1)$. Note that statics for the random scoring model is included in Appendix~\ref{Appendix:StatisticsRandom}.
\textbf{USAD} (\cite{audibert2020usad})
is based on adversely trained auto-encoders
in order to conduct unsupervised anomaly detection for multivariate time series.
\textbf{GDN} (\cite{deng2021graph}) is a structure learning approach based on graph neural networks for multivariate time series.
\textbf{AnomalyTrans} (\cite{xu2022anomaly}) proposes a new association-based criterion, which is embodied by a co-design of temporal models for learning more informative time point associations.

\section{Results}
\label{sec:results}

In Sec.~\ref{sec:PAdf}, we theoretically verify that the over-estimating problem of the PA protocol is solved through the mathematical formulation of the PAdf protocol. In this section, Actual possible problems of the PA protocol and the PA\%K protocol appearing in the examples of anomaly detection in Sec.~\ref{subsec:toy_result} and Sec.~\ref{subsec:toy} are explained. We also show that these problems are solved through the PAdf protocol. Additionally, in Sec.~\ref{subsec:bench_result}, the anomaly detection results of SOTA algorithms with the open dataset are analyzed and the advantages of evaluation with the PAdf protocol are explained. Note that more experiments are included in Appendix~\ref{Appendix:moreExp}. 

\subsection{Toy data}
\label{subsec:toy_result}

The evaluation result with respect to the data in Sec.~\ref{subsec:toy} is shown in Table~\ref{table:PAvsPAKvsPADF}.
In order to conduct a performance evaluation that satisfies \ref{item:1}, \ref{item:2}, and \ref{item:3} which are requirements of anomaly detector, the following conditions must be satisfied:
\begin{itemize}
  \item The F1 score of Fig.~\ref{fig:anomaly_case4} and Fig.~\ref{fig:anomaly_case6} should be higher than that of Fig.~\ref{fig:best}.
  \item The F1 score of Fig.~\ref{fig:anomaly_case8} should be lower than that of Fig.~\ref{fig:best}, Fig.~\ref{fig:anomaly_case4}, and Fig.~\ref{fig:anomaly_case6}.
  \item Fig.~\ref{fig:anomaly_case4} and Fig.~\ref{fig:anomaly_case6} may receive the same score from a monitoring point of view.
  \item For the performance evaluation satisfying \ref{item:3}, Fig.~\ref{fig:best} must score higher than Fig.~\ref{fig:worst}.
\end{itemize}

In the case of the PA protocol,
since the four cases in Fig.~\ref{fig:best}$\sim$\ref{fig:anomaly_case8} are given the same highest score, it is not possible to distinguish which detection performed the best.

The PA\%K protocol evaluates the highest score for the cases in \ref{fig:anomaly_case8}, where the detection is too late and causes losses that occurred as abnormal situations such as server failures or cyberattacks continued.
However, the PA\%K protocol gives the lowest scores for the cases of Fig.~\ref{fig:best} and \ref{fig:anomaly_case4} where detection is quick and accurate.

In the case of the PAdf protocol, 
Fig.~\ref{fig:anomaly_case4} and Fig.~\ref{fig:anomaly_case6}, which are the most ideal situations according to the requirements, are evaluated with the highest score of 0.933 with no under-estimation phenomenon. Also, Fig.~\ref{fig:anomaly_case8}, which has the highest number of detection  in the anomaly section but the detection is late, is given the lowest score. 
Moreover, Fig.~\ref{fig:best} is given a higher score than Fig.~\ref{fig:worst}.
Additionally, we verify the robustness with respect to the evaluations of the PAdf protocol in Appendix.~\ref{Appendix:robustness}.

\subsection{Benchmark data}
\label{subsec:bench_result}

For the {\bf MSL dataset} in Table~\ref{table:benchmark}, AnomalyTrans shows the best performance in the F1 score of the PA protocol, but it can be seen that the Random score also receives a high score of 0.907. This is a problem caused by the PA over-estimating all evaluations. The over-estimating problem seems to be solved to some extent through the F1 score calculated by the PA\%K protocol. For example, the score of AnomalyTrans, which is 0.943 when using the PA protocol, is lowered to 0.190 when using the PA\%K protocol. It can be seen that AnomalyTrans is overestimated by the PA protocol, while the values of other algorithms are reduced by half. It can be interpreted that this phenomenon occurs because AnomalyTrans finds anomaly segments in the MSL dataset well, but the rate of finding anomalies within each segment is low. However, in the PA\%K protocol, the performance of Random Score, USAD, and GDN does not differ significantly, and finding many anomalies within one segment is not important in an actual monitoring system. In the F1 score calculated using the ${\rm PAdf}_{0.7}$ protocol, there is a significant score difference between Random Score, USAD, and GDN. In this case, USAD obtains the highest score, 0.442, and AnomalyTrans shows a lower score value than Random Score. This shows that AnomalyTrans has the highest F1 score evaluated through the PA protocol, but the lowest F1 score evaluated through the ${\rm PAdf}_{0.7}$ protocol as shown in Table~\ref{table:benchmark}. Also, for AnomalyTrans, the ratio of ${\rm PAdf}_{0.9}$ to ${\rm PAdf}_{0.7}$ is 45.9\% as shown in Table~\ref{table:ratioofPAdf}. The above results show that although AnomalyTrans finds a large number of anomaly segments in the MSL dataset, the PAdf score is lowered by the large gap between the occurrence and detection of the anomaly situation.

For the {\bf SMAP dataset}, AnomalyTrans records the best score by the PA protocol.
However, Random Score and USAD receive similar F1 scores of PA protocols. 
Also, GDN is evaluated to perform 82\% of Random Score. 
The F1 score of the ${\rm PAdf}_{0.7}$ protocol is 78\% for USAD, 58\% for GDN, and 84\% for AnomalyTrans compared to Random Score. The large difference in the performance of the model compared to that evaluated by the PA protocol can be seen as a problem caused by the overestimation of the PA protocol. According to the evaluation method of ${\rm PAdf}_{0.7}$, the anomaly detection performance of the SOTA models is all lower than that of the Random Score, indicating that the models do not perform well on the SMAP dataset. However, according to Table~\ref{table:ratioofPAdf}, USAD has the highest ratio of ${\rm PAdf}_{0.9}$ and ${\rm PAdf}_{0.7}$ with 98.8\%.From this, it can be inferred that the time between the occurrence and detection of an anomaly segment is the shortest in the case of USAD.

In the {\bf SWaT dataset}, AnomalyTrans receives an overwhelmingly high evaluation when evaluated by the PA protocol. However, since AnomalyTrans shows the lowest performance with PA\%K and ${\rm PAdf}_{0.7}$, it can be seen that the model performance is overestimated by the PA protocol. Also, according to the F1 score of the ${\rm PAdf}_{0.7}$ protocol, the scores of the SOTA models are lower than the Random Score. It suggests that hyperparameter tuning or improvement of anomaly detection models is needed.

Summarizing the above results, AnomalyTrans records the highest scores in all datasets with the F1 scores of the PA protocol, but most of the lowest scores in all datasets with the F1 score of ${\rm PAdf}_{0.7}$. On the other hand, USAD shows the opposite result to AnomalyTrans.
The difference in scores with PA and PAdf depends on whether or not the anomaly detection speed is considered.
From the result of Table~\ref{table:ratioofPAdf}, the fact that USAD has a higher ratio than that of AnomalyTrans indicates that USAD detects anomaly segments faster than AnomalyTrans. According to the results, it can be seen that USAD is the best anomaly detection model among the comparative models when comprehensively considering both the number of detected anomalies and the speed of the anomaly detection.

The performance of SOTA models obtained through the PA protocol is generally evaluated to be higher than the Random Score.
However, when the PAdf protocol is used, all SOTA models are evaluated to perform worse than the Random Score model except for the MSL dataset. This is because the performance of current SOTA models is developed according to the PA protocol. 
Therefore, the performance of detecting anomaly segments is good, but the performance of quickly discovering anomaly segments, which is one of the important elements in the real-world monitoring system, is not good. 
In the future, in the field of time-series anomaly detection, it is expected that detection models applicable to the real world will be developed in the direction of quickly finding many segments without delay if the PAdf protocol is used as an evaluation metric for anomaly detection models.

\begin{table}[t]
\caption{The ratio of F1 score using ${\rm PAdf}_{0.7}$ to F1 score using ${\rm PAdf}_{0.9}$.}
\label{table:ratioofPAdf}
\vskip 0.15in
\begin{center}
\begin{small}
\begin{sc}
\begin{tabular}{lccc}
\toprule
model & MSL & SMAP & SWaT \\
\midrule
USAD & 93.4\% & 98.8\% & 100\% \\
GDN  & 77.9\%  & 72.6\% & 100\% \\
AnomalyTrans    & 45.9\% & 74.1\% & 78.9\% \\
\bottomrule
\end{tabular}
\end{sc}
\end{small}
\end{center}
\vskip -0.1in
\end{table}

\section{Conclusion and Discussion}
\label{sec:sum}

In this paper, the anomaly detection evaluation metric with a decay function called PAdf is introduced in the anomaly detection monitoring system with mathematical formulations. The PAdf protocol satisfies the three most important requirements \ref{item:1}$\sim$\ref{item:3}, and its evaluations have been verified theoretically and experimentally. The PAdf protocol solves the over- and under-estimation problems which are occurred in the existing scoring protocol such as the PA and PA\%K. 

The PA protocol only focuses on finding many anomalous segments, 
whereas the score of the PAdf protocol considers not only finding many segments but also detecting anomalies quickly without delay.
So, the PAdf protocol plays an important role in finding a time-series anomaly detection algorithm that is useful for real-world monitoring systems and managing performance in our current applications such as AD exchange monitoring system.



We note that the formula derived in this paper assumes the case where there is only one anomaly segment ${\cal A}$. In the case of many anomaly segments, generalized metrics can be obtained by adding summation for the segments. But, in the released code, precision, recall, and F1 score are calculated through the PAdf protocol from the ground truth and the scores without restrictions on the number of anomaly segments.

A potential limitation of the proposed method is the need to determine a decay function or decay rate for evaluation. Although it is recommended to set the default value of the decay rate $d$ to 0.9 and adjust the value according to the loss caused by late anomaly detection, it is necessary to determine parameters that can sometimes feel ambiguous for evaluation. 

As a future work, 
we will evaluate the time-series anomaly detection model using the PAdf protocol with 
the techniques including an extension of traditional AUC-ROC and AUC-PR measures and introducing a continuous label to enable more flexibility in measuring detected anomaly ranges, which are used in \cite{paparrizos2022volume}.



% ...





\nocite{langley00}
\bibliographystyle{unsrtnat}
\bibliography{neurlps_2023}

\newpage
\appendix
\onecolumn
\section{Reproducibility}
\label{Appendix:Reproducibility}
We append the code to reproduce our results.
Which code reproduces which result is described as follows:
\begin{itemize}
  \item {\bf Figure3.ipynb} gives the graphs in Fig.~\ref{fig:case1}.
  \item {\bf Table1-various\_case.ipynb} is the code to reproduce the F1 scores with respect to PA, PA\%K, PAdf as shown in  Table.~\ref{table:PAvsPAKvsPADF}.
  \item {\bf dataset/} directory has the ground truths of each dataset and the prediction results from the SOTA models, which save in the format of a pickle. We notice that the prediction results of each algorithm may be different depending on pre-processing or fine-tuning even though the same dataset and the same model are used.
  \item {\bf evaluation.py} is the code to evaluate the prediction results in {\bf dataset/}. Then, the scores in Table.~\ref{table:benchmark} can be reproduced. In the file, how to operate each case is written as a comment.
\end{itemize}

\section{Baselines}
\label{Appendix:Baselines}

\textbf{Random score} is randomly generated from  a uniform
distribution such as  $p(t) \sim U(0,1)$.
The reason for using the random score model is to study the overestimation phenomenon from the PA protocol with the benchmark data.
In addition, although the random score model is not an appropriate control group in the AB test, it can also be used for the purpose of interpreting that the model is meaningful only when the score is at least higher than that of the random score model.
Since there is a possibility of cherry-picking by producing results several times due to randomness, the results in Table~\ref{table:benchmark} are obtained by averaging 5 experimental results.

\textbf{USAD} (\cite{audibert2020usad})
is based on adversely trained auto-encoders
in order to conduct unsupervised anomaly detection for multivariate
time series.
By adopting adversarial training and its architecture, this algorithm has the advantages of fast learning and stable anomaly detection.

\textbf{GDN} (\cite{deng2021graph}) is a structure learning approach based on graph neural networks for multivariate time series.
This algorithm has the advantage of being able to provide interpretability for detected anomalies.

\textbf{AnomalyTrans} (\cite{xu2022anomaly}) proposes a new association-based criterion, which is embodied by a co-design of temporal models for learning more informative time point associations. 
This algorithm includes anomaly attention with two branch structures to embody this association discrepancy and uses the min-max strategy to amplify the difference between normal and abnormal points.

Note that the results in Table~\ref{table:benchmark} are reproduced.
When we conduct the experiments,
the open codes in GitHub of each research are used with the default settings and no other pre- or post-processings and fine-tunings are applied.
Each score was calculated from the same predicted values and ground truth, only with a different protocol: PA, PA\%K, PAdf.
Thresholds are determined as the value giving the best score according to the evaluation methodology of previous research (\cite{xu2022anomaly, kim2022towards}).

\section{F1 scores of the PAdf protocol in the various and specific cases}
\label{Appendix:mathVariousPAdf}

In this section, we analytically derive F1 score by the PAdf protocol in the case of various probability distributions.
Recall that the general forms of recall  \eqref{eq:RPAdf1} and precision  \eqref{eq:PPAdf1} are given by
\begin{align}
\mathbb{R}_{\rm PAdf} &= \sum^{N-1}_{n=0}{\cal D}(t_{s+n}) (1-p(t_{s+n})) \prod^{n-1}_{i=0} p(t_{s+i}), \notag\\
\mathbb{P}_{\rm PAdf} &= \frac{{\cal P}_{\cal A} \cdot \mathbb{R}_{\rm PAdf}}{\tau^{-1}\sum^{\tau-N}_{t\in {\cal A}^{\rm c}}(1-p(t))+\left(1-\prod^N_{i=0}p(t_i)\right){\cal P}_{\cal A}}. \notag
\end{align}
From now on, we will assume the specific case that the detection algorithm gives the detection probability 1 or 0 for simplicity.

\subsection{only $p(t_s)=0$, otherwise $p(t)=1$}
In all other cases, the algorithm is considered normal.
Following the third requirement \ref{item:3},
this case is the best detection.
This is the best result the detection algorithm can give.
Then, Eq.~\eqref{eq:RPAdf1} and  Eq.~\eqref{eq:PPAdf1} are rewritten as
\begin{align}
\mathbb{R}_{\rm PAdf} &= {\cal D}(t_{s}) (1-p(t_{s})) + \sum^{N-1}_{n=1}{\cal D}(t_{s+n}) (1-p(t_{s+n})) \prod^{n-1}_{i=0} p(t_{s+i}) = {\cal D}(t_{s}) = 1, \\
\mathbb{P}_{\rm PAdf} &= \frac{{\cal P}_{\cal A}}{{\cal P}_{\cal A}} = 1.
\end{align}
Finally, we can obtain the F1 score as
\begin{align}
    \mathbb{F}1_{\rm PAdf} = 1.
\end{align}
The highest score is given by the PAdf protocol in this case as we guess.

\subsection{only $p(t_e)=0$, otherwise $p(t)=1$}
In this case, the detection algorithm detects only one data point at $t=t_e$, but this is the end point of the anomaly segment.
From Eq.~\eqref{eq:RPAdf1} and  Eq.~\eqref{eq:PPAdf1},
we can derive the recall and precision as follows:
\begin{align}
\mathbb{R}_{\rm PAdf} &= \sum^{N-2}_{n=0}{\cal D}(t_{s+n}) (1-p(t_{s+n})) \prod^{n-1}_{i=0} p(t_{s+i}) + {\cal D}(t_{s+(N-1)}) (1-p(t_{s+N-1})) \prod^{N}_{i=0} p(t_{s+N-1}), \\
&= {\cal D}(t_{s+(N-1)}) (1-p(t_{s+N-1})) \prod^{N}_{i=0} p(t_{s+N-1}), \\
&= {\cal D}(t_{s+(N-1)}) \prod^{N}_{i=0} 1 \\
&= {\cal D}(t_e) \\
\mathbb{P}_{\rm PAdf} &= \frac{{\cal P}_{\cal A} {\cal D}(t_{s+(N-1)})}{\tau^{-1}\sum^{\tau-N}_{t\in {\cal A}^{\rm c}}1+{\cal P}_{\cal A}} 
= \frac{{\cal P}_{\cal A} {\cal D}(t_{s+(N-1)})}{{\cal P}_{\cal A}} 
= {\cal D}(t_{e}),
\end{align}
where ${\cal P}_{\cal A} = N/\tau$ and $t_{s+N-1} = t_e$.
Then, the F1 score is 
\begin{align}
    \mathbb{F}1_{\rm PAdf} = {\cal D}(t_{e}).
\end{align}
If Eq.~\eqref{eq:decay_func} is assumed,
\begin{align}
    \mathbb{F}1_{\rm PAdf} = d^{N},
\end{align}
where $N$ is the length of the anomaly segment as
$N = t_{e}-t_s$.



\subsection{$p(t_{s+2})=p(t_{s+5})=0$, $\{t_{s+2}, t_{s+5}\} \subset {\cal A}$, and $p(t_{20})=p(t_{25})=1$, $\{ t_{s+20}, t_{s+25} \} \subset {\cal A}^{\rm c}$, otherwise $p(t)=1$}
Let us calculate the F1 score of the PAdf protocol in the complicated case.
The algorithm detects the anomalies at $t=t_{s+2}$ and $t=t_{s+5}$, which contains the segment of contiguous anomalies.
Additionally, there are false positives at $t=t_{s+20}$ and $t=t_{s+25}$.
Then,
The general forms of recall \eqref{eq:RPAdf1} and  precision \eqref{eq:PPAdf1} are
rewritten as 
\begin{align}
\mathbb{R}_{\rm PAdf} &= {\cal D}(t_{s+2}) (1-p(t_{s+2})) \prod^{1}_{i=0} p(t_{s+i}) 
+ {\cal D}(t_{s+5}) (1-p(t_{s+5})) \prod^{4}_{j=0} p(t_{s+j}), \\
&= {\cal D}(t_{s+2}) \\
\mathbb{P}_{\rm PAdf} &= \frac{{\cal P}_{\cal A} {\cal D}(t_{s+2})}{2\tau^{-1}+{\cal P}_{\cal A}}
= \frac{N}{2+N}{\cal D}(t_{s+2}),
\end{align}
where the false positives only affect the precision.
Finally, the F1 scores are obtained as
\begin{align}
    \mathbb{F}1_{\rm PAdf} = \frac{N}{1+N} {\cal D}(t_{s+2}).
\end{align}
Here, we note that the F1 score of the PAdf protocol in the case of $M$ false positives is derived as
\begin{align}
    \mathbb{F}1_{\rm PAdf} = \frac{2N}{2N+M} {\cal D}(t_{s+2}),
\end{align}
where the F1 score becomes zero as the number of false positive $M$ goes to infinity.
In addition, the F1 score is equal to ${\cal D}(t_{s+2})$ if there is no false positive.

\section{Robustness}
\label{Appendix:robustness}

Previously, in the introduction, we define the requirements for the ideal time series anomaly detection model as follows:
\begin{enumerate}[label=\textbf{R.\arabic*}]
    \item \label{item:1} {\bf The abnormal situations should be detected without missing them.} 
    \item \label{item:2} {\bf The abnormal situations should be detected as soon as possible.}
    \item \label{item:3} {\bf The frequency of false alarms should be low.}
\end{enumerate}

\begin{figure}[t] 
  \begin{center}
\subfigure[{~Anomaly detection case 1}]{
\includegraphics[width=0.46\textwidth]{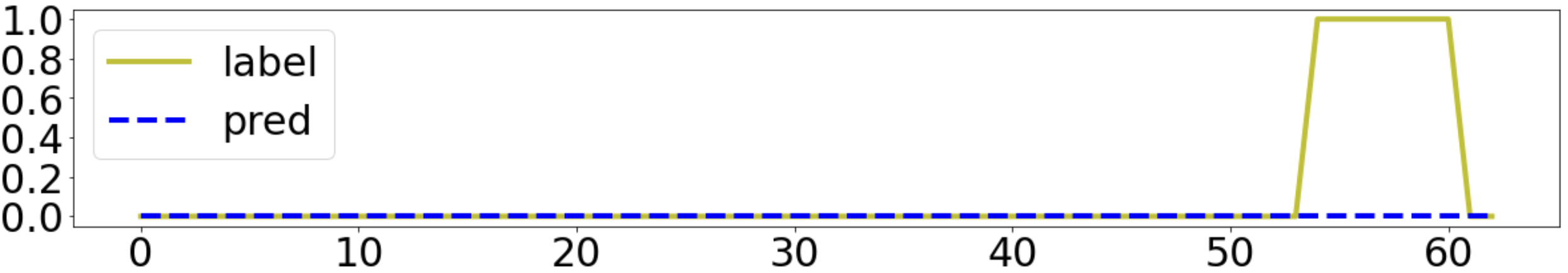}\label{fig:A_case_0}} 
\subfigure[{~Anomaly detection case 2}]{
\includegraphics[width=0.46\textwidth]{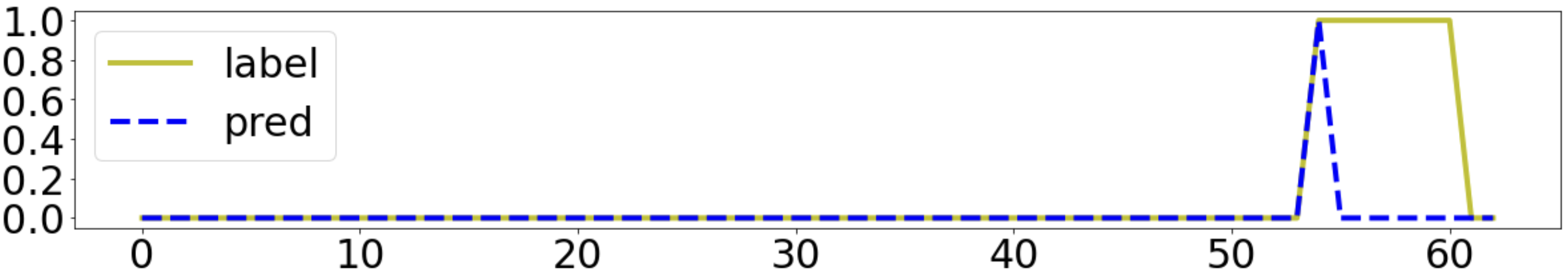}\label{fig:A_case_1}} \\
\subfigure[{~Anomaly detection case 3}]{
\includegraphics[width=0.46\textwidth]{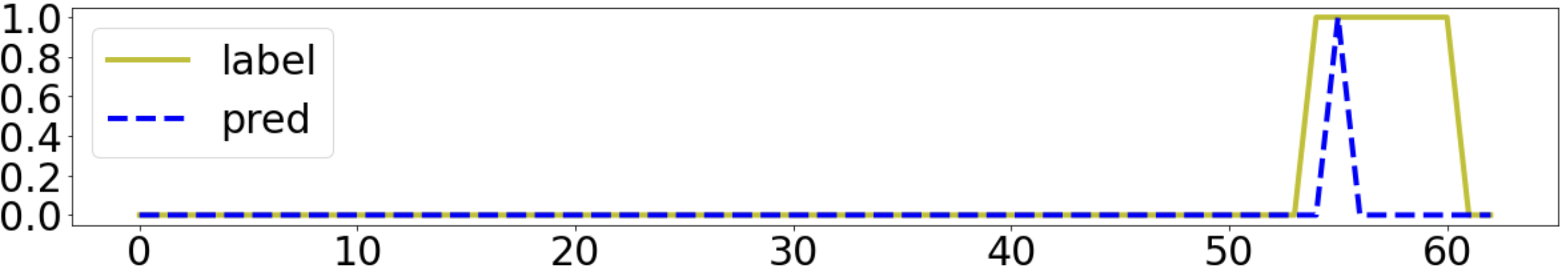}\label{fig:A_case_2}}  
\subfigure[{~Anomaly detection case 4}]{
\includegraphics[width=0.46\textwidth]{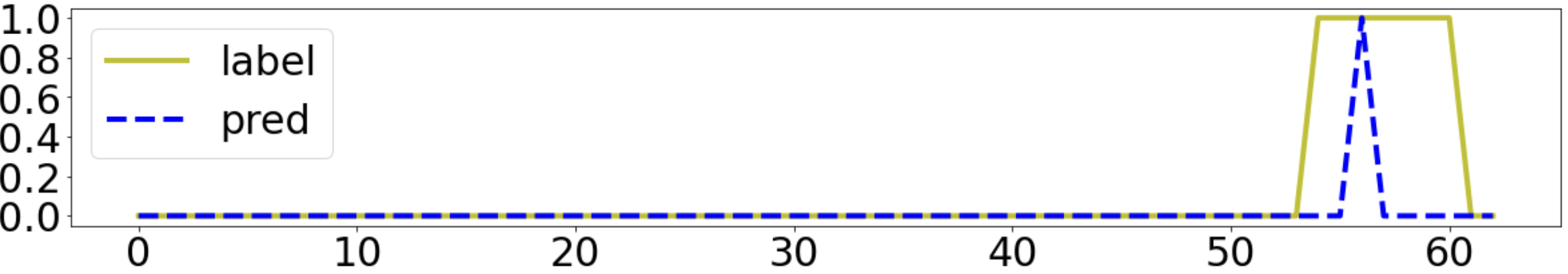}\label{fig:A_case_3}} \\
\subfigure[{~Anomaly detection case 5}]{
\includegraphics[width=0.46\textwidth]{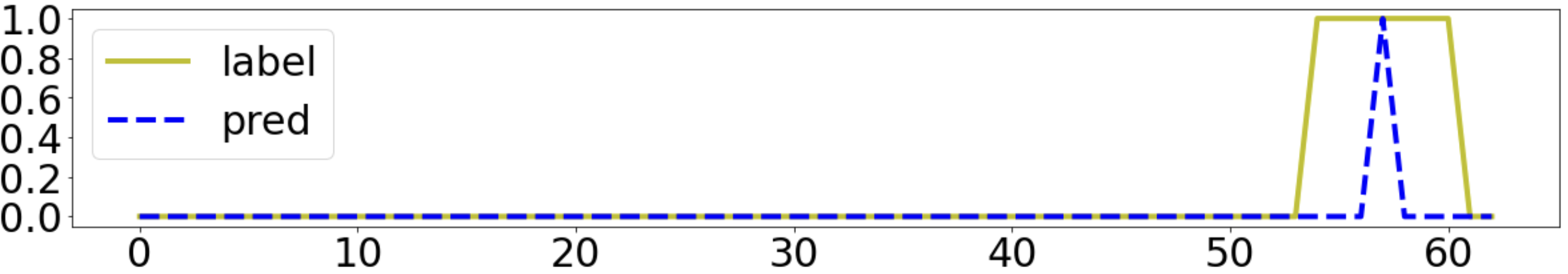}\label{fig:A_case_4}} 
\subfigure[{~Anomaly detection case 6}]{
\includegraphics[width=0.46\textwidth]{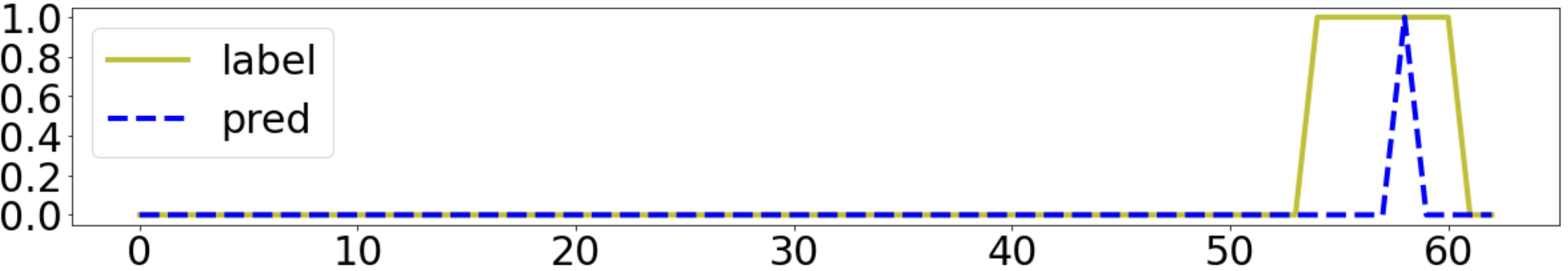}\label{fig:A_case_5}} \\
\subfigure[{~Anomaly detection case 7}]{
\includegraphics[width=0.46\textwidth]{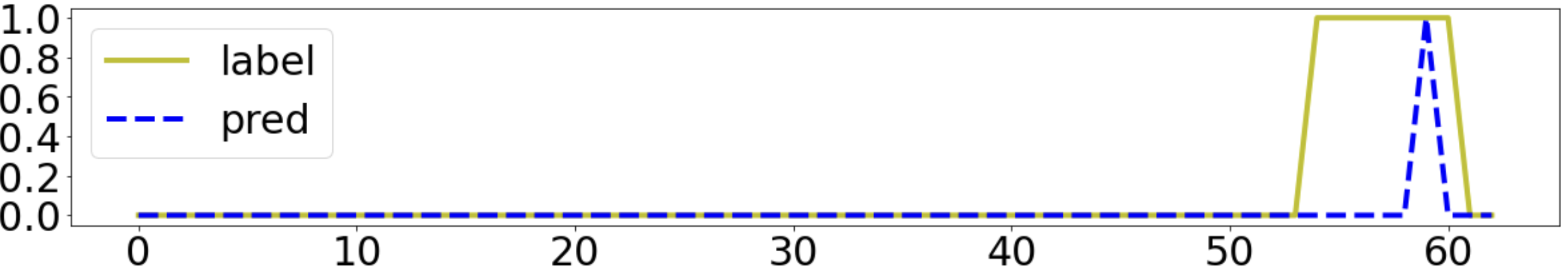}\label{fig:A_case_6}} 
\subfigure[{~Anomaly detection case 8}]{
\includegraphics[width=0.46\textwidth]{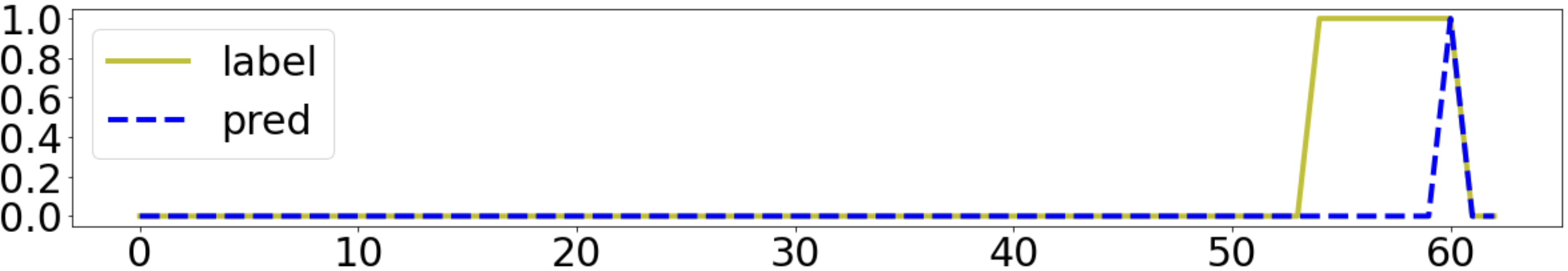}\label{fig:A_case_7}} \\
\subfigure[{~Anomaly detection case 9}]{
\includegraphics[width=0.46\textwidth]{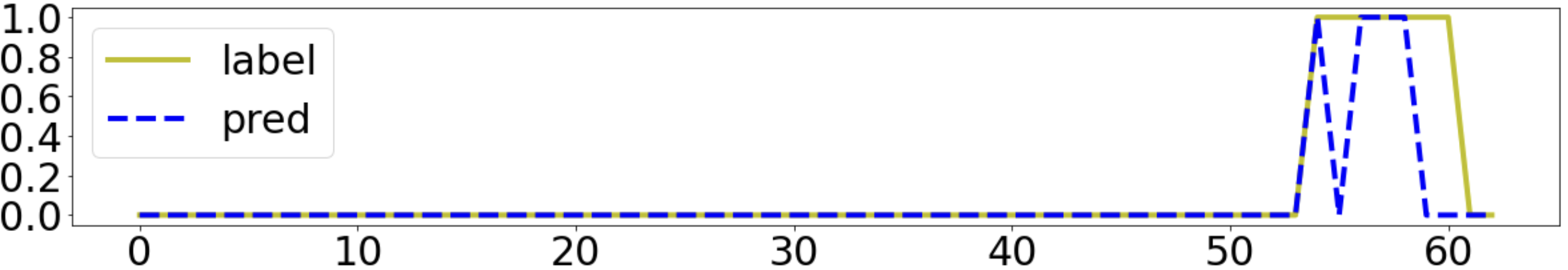}\label{fig:A_case_8}} 
\subfigure[{~Anomaly detection case 10}]{
\includegraphics[width=0.46\textwidth]{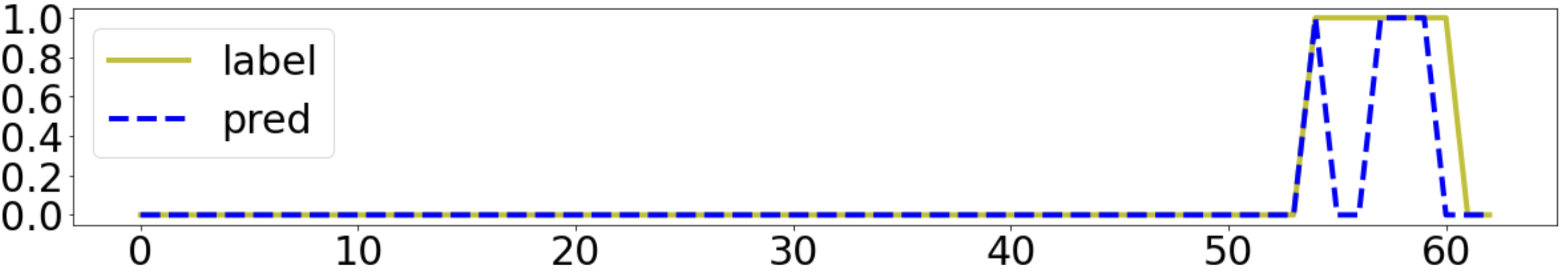}\label{fig:A_case_9}} \\
\subfigure[{~Anomaly detection case 11}]{
\includegraphics[width=0.46\textwidth]{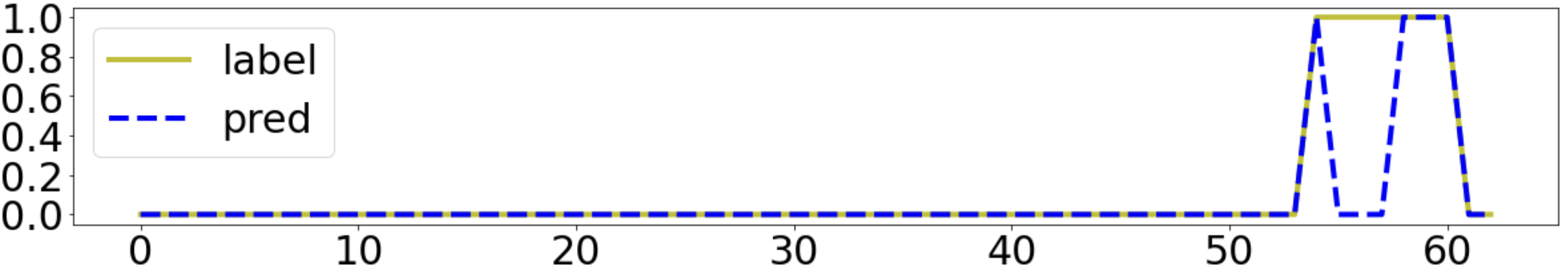}\label{fig:A_case_10}} 
\subfigure[{~Anomaly detection case 12}]{
\includegraphics[width=0.46\textwidth]{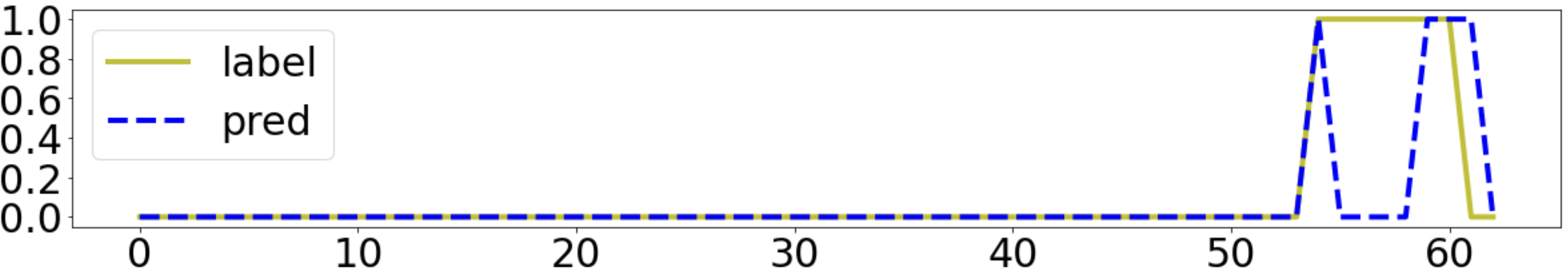}\label{fig:A_case_11}} \\
\subfigure[{~Anomaly detection case 13}]{
\includegraphics[width=0.46\textwidth]{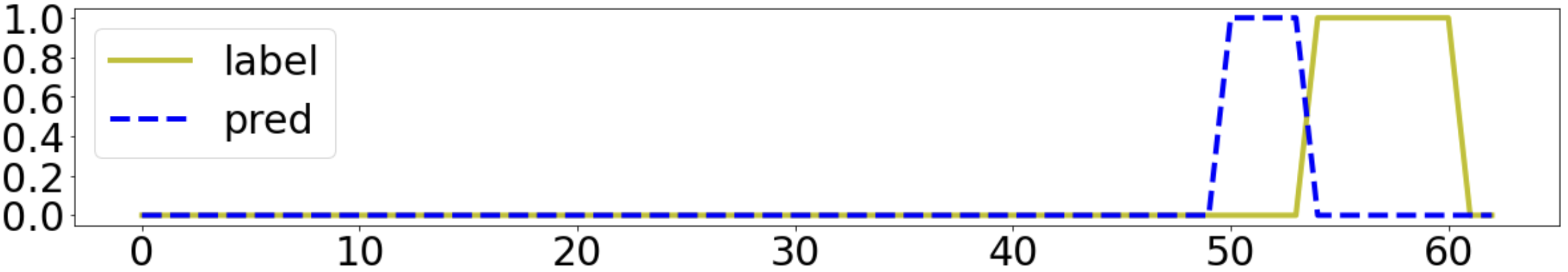}\label{fig:A_case_12}}
\subfigure[{~Anomaly detection case 14}]{
\includegraphics[width=0.46\textwidth]{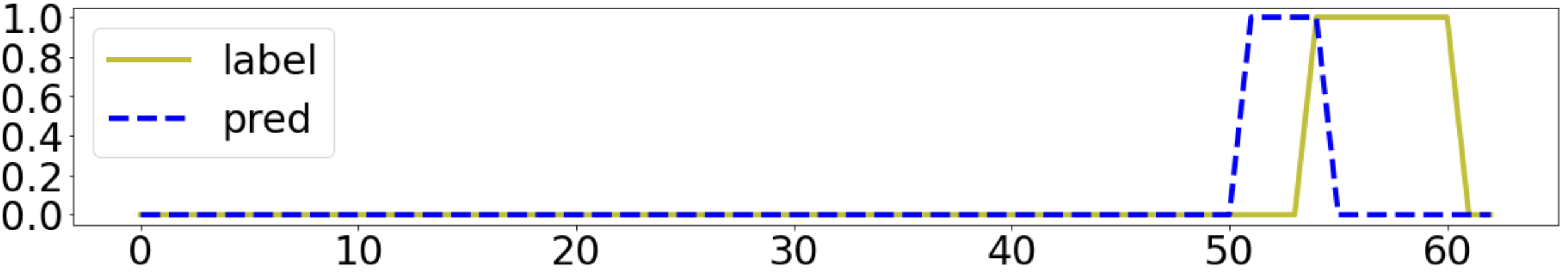}\label{fig:A_case_13}} \\
\subfigure[{~Anomaly detection case 15}]{
\includegraphics[width=0.46\textwidth]{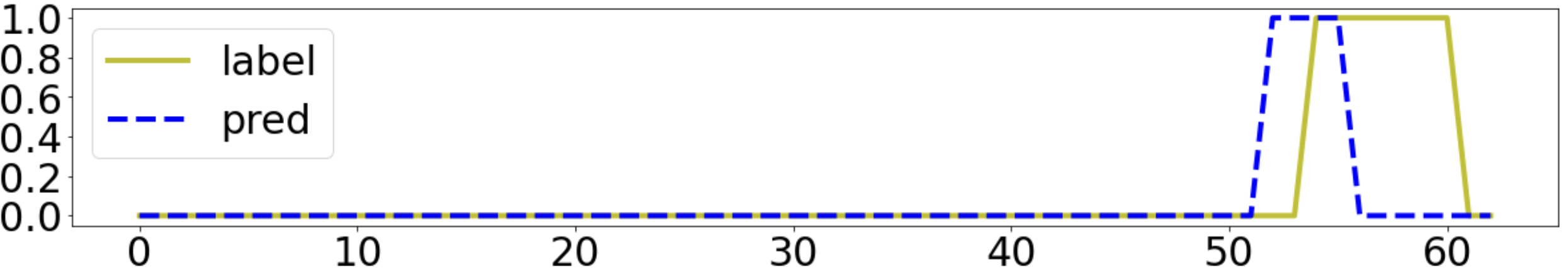}\label{fig:A_case_14}} 
\subfigure[{~Anomaly detection case 16}]{
\includegraphics[width=0.46\textwidth]{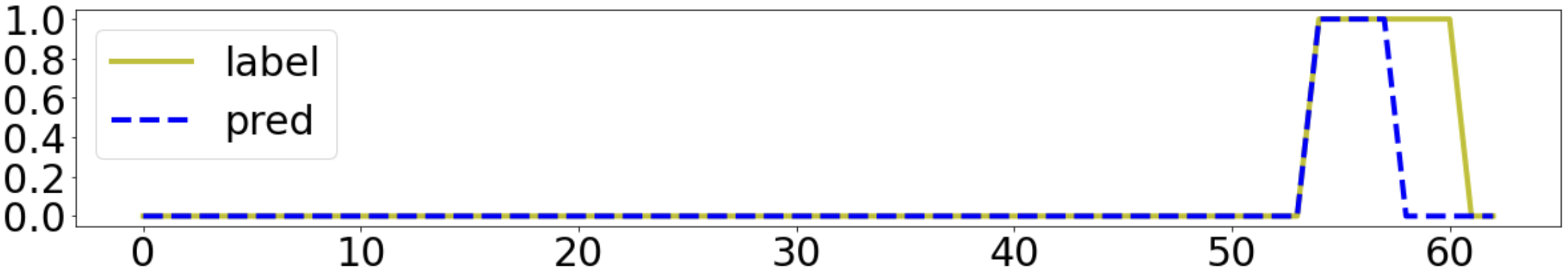}\label{fig:A_case_15}} 
  \end{center}
  \caption{various cases. The blue line denotes the signal from the detector, and the yellow line means the ground truth. } \label{fig:A_cases}
\end{figure}

In order to verify whether the F1 score of PAdf protocol evaluates performance robustly in accordance with the requirements in various situations, scores are obtained for multiple cases in Fig.~\ref{fig:A_cases}. 
Fig.~\ref{fig:A_case_0} is the case that all are detected as 0. 
Fig.~\ref{fig:A_case_1} to Fig.~\ref{fig:A_case_7} show the cases from the earliest to the latest detection within the anomaly segment when anomaly occurred, respectively.
In Fig.~\ref{fig:A_case_8} to Fig.~\ref{fig:A_case_11}, the first detection occurs immediately after the anomaly situation, but the subsequent detection is delayed.
Fig.~\ref{fig:A_case_12} is the case that an anomaly situation is continuously predicted, but the prediction is a false positive.
From Fig.~\ref{fig:A_case_13} to Fig.~\ref{fig:A_case_15}, the intersection between continuous detection results and the ground truth increases.
As can be seen in Table~\ref{table:A_case}, Fig.~\ref{fig:A_case_0} has a score of 0. From Fig.~\ref{fig:A_case_1} to Fig.~\ref{fig:A_case_7}, although anomaly segments are found, the detection is delayed, resulting in a continuous decrease in the score. 
In Fig.~\ref{fig:A_case_8} to Fig.~\ref{fig:A_case_10}, since the first detection accurately found the anomaly segment, the same score of 1.0 is given even if the subsequent detection is delayed.
In the case of Fig.~\ref{fig:A_case_11}, since the false positive is reflected in the score, the score is lower than 1.0.
In Fig.~\ref{fig:A_case_12}, a score of 0 is assigned because all are false positives.
From Fig.~\ref{fig:A_case_13} to Fig.~\ref{fig:A_case_15}, since the anomaly segment is detected as soon as it occurred, the score gets lower if the false positive area gets wider.
As such, it is confirmed that the F1-score of PAdf protocol robustly assigns scores according to the R1, R2, and R3 which are described as requirements for the ideal time series anomaly detection model earlier.

\begin{table}[t]
\caption{{\bf PAdf} with the decay rates $d=0.9$ is calculated for the cases in Fig.~\ref{fig:A_cases}.}
\label{table:A_case}
\vskip 0.15in
\begin{center}
\begin{small}
\begin{sc}
\begin{tabular}{lcccccccc}
\toprule
Figure & \ref{fig:A_case_0} & \ref{fig:A_case_1} & \ref{fig:A_case_2} & \ref{fig:A_case_3} &\ref{fig:A_case_4} &\ref{fig:A_case_5} &\ref{fig:A_case_6} &\ref{fig:A_case_7} \\
\hline
${\rm PAdf}_{0.9}$ & 0.0 & 1.0 & 0.95  & 0.90  & 0.84 & 0.79 & 0.74 & 0.69 \\
\midrule
Figure &\ref{fig:A_case_8} &\ref{fig:A_case_9} &\ref{fig:A_case_10} &\ref{fig:A_case_11} &\ref{fig:A_case_12} &\ref{fig:A_case_13} &\ref{fig:A_case_14} &\ref{fig:A_case_15}   \\
\hline
${\rm PAdf}_{0.9}$ & 1.0 & 1.0 & 1.0 & 0.93 & 0.0 & 0.82 & 0.875 & 1.0 \\
\bottomrule
\end{tabular}
\end{sc}
\end{small}
\end{center}
\vskip -0.1in
\end{table}


\section{Statistics for Random Scoring Models}
\label{Appendix:StatisticsRandom}

\begin{figure}[t] 
  \begin{center}
\subfigure[{~MSL}]{
\includegraphics[width=0.46\textwidth]{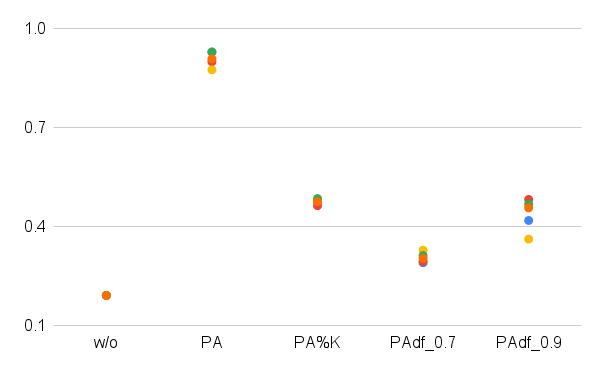}\label{fig:MLS_rand_var}} 
\subfigure[{~SMAP}]{
\includegraphics[width=0.46\textwidth]{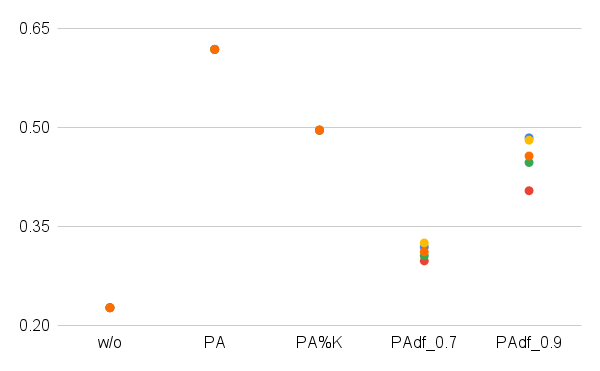}\label{fig:SMAP_rand_var}}
\subfigure[{~SWaT}]{
\includegraphics[width=0.46\textwidth]{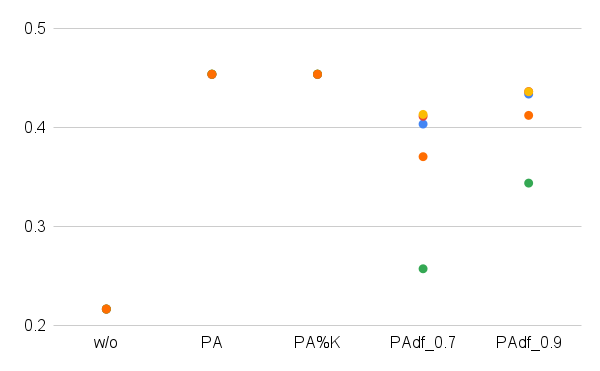}\label{fig:SWaT_rand_var}}  
  \end{center}
  \caption{This Figure shows the distributions of the F1 score without any protocols, that of the PA, that of the PA\%K, and that of the PAdf.} \label{fig:rand_var}
\end{figure}

As shown in Table~\ref{table:benchmark}, the value of the random score model is the average value of detection results obtained by 5 random samplings from a uniform distribution.
In Table~\ref{table:rand_var}, the mean and variance of the score are calculated for each benchmark dataset. According to the visualization in Fig.~\ref{fig:rand_var}, the variance shows that the existing metrics do not show randomness in the model results, but the score distribution of the PAdf protocol shows the score distribution by random score model.

\begin{table*}[h]
\caption{Average and variance of the random score model shown in Table~\ref{table:benchmark}.}\label{table:rand_var}
\vskip 0.15in
\begin{center}
\begin{sc}
\begin{tabular}{llccccc}
\toprule
dataset & {} & w/o & PA & PA\%K & ${\rm PAdf}_{0.7}$ &  ${\rm PAdf}_{0.9}$ \\
\midrule
MSL & average & 0.191 & 0.907 & 0.475 & 0.306 &  0.437 \\ %
{} & variance & 7.6e-09 & 5.1e-04 & 6.1e-05 & 2.3e-04 &  2.3e-03 \\ %
\hline
SMAP & average 
& 0.227 & 0.618 & 0.496 & 0.312 &  0.454 \\ 
{} & variance & 6.4e-11  & 1.5e-08 & 1.9e-08 & 1.1e-04 &  1.0e-03 \\ %
\hline
SWaT & average 
& 0.217 & 0.453 & 0.453 & 0.371 & 0.412 \\ 
{} & variance & 4.8e-10 & 3.8e-08 & 3.7e-08 & 4.3e-03 &  1.6e-03 \\ %
\bottomrule
\end{tabular}
\end{sc}
\end{center}
\vskip -0.1in
\end{table*}

Table~\ref{table:benchmark} is the F1 score of each protocol obtained using the SOTA anomaly detection models and the benchmark dataset introduced in Sec.~\ref{subsec:benchmark}. As mentioned earlier, since we train and test the SOTA model for each benchmark data without any hyperparameter tuning, the score may change depending on the degree of fine-tuning of the model. Table~\ref{table:ratioofPAdf} is the ratio of the F1 score of ${\rm PAdf}_{0.7}$ to the F1 score of ${\rm PAdf}_{0.9}$. It is calculated for each model and dataset. This ratio is a numerical value that shows how fast an anomaly segment is found.


\section{More Experiments with TSB-UAD pipeline}
\label{Appendix:moreExp}

\subsection{HEX/UCR Datasets}

\begin{figure}[t] 
  \begin{center}
\subfigure[{~NOISEBIDMC1\_2500 dataset and IForest model }]{
\includegraphics[width=0.99\textwidth]{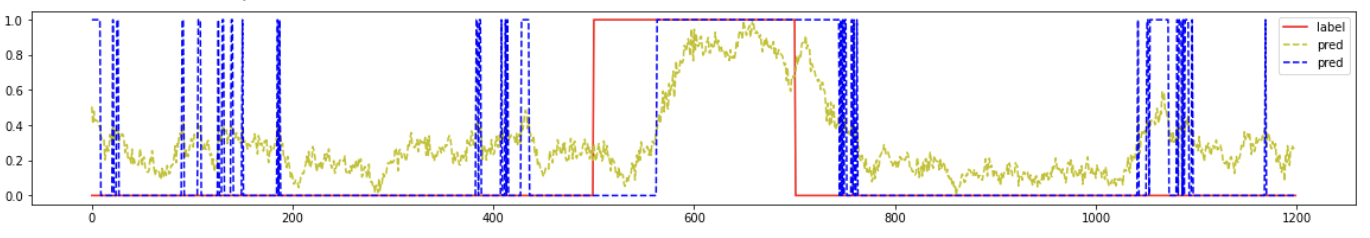}\label{fig:2500-IForest}} 
  \end{center}
  \caption{This figure shows the sample case in Table~\ref{table:HEXUCR}} \label{fig:samplecase}
\end{figure}

Additional experiments are conducted on 5 datasets among the HEX/UCR TSAD datasets (detailed in \cite{UCRArchive2018}). Since they are the univariate time-series data, we used the TSB-UAD pipeline and the models which the pipeline has.
Interestingly, in the case of the NOISEBIDMC1 dataset, the IForest model wins OCSVM and the random score models by the score of PA, but the OCSVM and the random score models win the IForest model by the score of PAdf. This is because the detection of the IForest is delayed as shown in Fig.~\ref{fig:samplecase}.

\begin{table*}[h]
\caption{HEX/UCR Datasets (\cite{UCRArchive2018}): {\bf PAdf} is calculated with two decay rates as $d=0.9, 0.99$ which are denoted as 
${\rm PAdf}_{0.9}$ and ${\rm PAdf}_{0.99}$, respectively. 
The parameters $K$ of PA\%K are fixed as $K=20$. Note that {\bf w/o} denotes the F1 score without any protocols.}\label{table:HEXUCR}
\vskip 0.15in
\begin{center}
\begin{sc}
\begin{tabular}{llccccc}
\toprule
dataset & model & w/o & PA & PA\%K & ${\rm PAdf}_{0.9}$ &  ${\rm PAdf}_{0.99}$ \\
\midrule
NOISEBIDMC1
& SAND & 0.622 & 0.206 & 0.206 & 0.057 & 0.161 \\
{} & PCA & 0.348 & 0.160 & 0.160 & 0.070 & 0.070 \\
{} & OCSVM & 0.138 & 0.162 & 0.162 & 0.162 & 0.162 \\
{} & MatrixProfile & 0.524 & 0.194 & 0.194 & 0.073 & 0.110 \\
{} & LOF & 0.377 & 0.166 & 0.166 & 0.040 & 0.104 \\
{} & IForest & 0.549 & 0.177 & 0.177 & 0.090 & 0.098 \\
{} & Random Score & 0.039 & 0.158 & 0.121 & 0.120 & 0.152 \\
\hline
GP711Marker & SAND & 0.937 & 1.000 & 1.000 & 1.000 & 1.000 \\
LFM5z4 & PCA & 0.074 & 0.074 & 0.074 & 0.074 & 0.074 \\
{} & OCSVM & 0.037 & 0.103 & 0.050 & 0.026 & 0.044 \\
{} & MatrixProfile & 0.835 & 0.832 & 0.832 & 0.781 & 0.827 \\
{} & LOF & 0.025 & 0.025 & 0.025 & 0.025 & 0.025 \\
{} & IForest & 0.391 & 0.121 & 0.121 & 0.121 & 0.121 \\
{} & Random Score & 0.028 & 0.095 & 0.077 & 0.057 & 0.088 \\
\hline
weallwalk
& SAND & 0.012 & 0.029 & 0.029 & 0.029 & 0.029 \\
{} & PCA & 0.007 & 0.030 & 0.020 & 0.007 & 0.023 \\
{} & OCSVM & 0.163 & 0.033 & 0.033 & 0.033 & 0.033 \\
{} & MatrixProfile & 0.103 & 0.033 & 0.033 & 0.033 & 0.033 \\
{} & LOF & 0.055 & 0.042 & 0.042 & 0.022 & 0.039 \\
{} & IForest & 0.026 & 0.032 & 0.032 & 0.007 & 0.026 \\
{} & Random Score & 0.025 & 0.035 & 0.025 & 0.032 & 0.035 \\
\hline
ECG2
& SAND & 0.096 & 0.096 & 0.096 & 0.087 & 0.095 \\
{} & PCA & 0.305 & 0.167 & 0.167 & 0.062 & 0.152 \\
{} & OCSVM & 0.007 & 0.040 & 0.007 & 0.013 & 0.036 \\
{} & MatrixProfile & 0.521 & 0.616 & 0.616 & 0.321 & 0.488 \\
{} & LOF & 0.770 & 0.754 & 0.754 & 0.561 & 0.705 \\
{} & IForest & 0.057 & 0.099 & 0.099 & 0.034 & 0.085 \\
{} & Random Score & 0.011 & 0.098 & 0.032 & 0.055 & 0.092 \\
\hline
DISTORTEDGP711 & SAND & 0.936 & 1.000 & 1.000 & 1.000 & 1.000 \\
MarkerLFM5z3 & PCA & 0.185 & 0.224 & 0.209 & 0.220 & 0.220 \\
{} & OCSVM & 0.500 & 0.868 & 0.713 & 0.713 & 0.847 \\
{} & MatrixProfile & 0.692 & 0.868 & 0.807 & 0.713 & 0.776 \\
{} & LOF & 0.633 & 0.742 & 0.713 & 0.652 & 0.666 \\
{} & IForest & 0.055 & 0.056 & 0.056 & 0.056 & 0.056 \\
{} & Random Score & 0.017 & 0.044 & 0.030 & 0.028 & 0.042 \\
\bottomrule
\end{tabular}
\end{sc}
\end{center}
\vskip -0.1in
\end{table*}


\subsection{More Datasets}

\begin{figure}[t] 
  \begin{center}
\subfigure[{~${\rm PAdf}_{0.9}$ and w/o}]{
\includegraphics[width=0.8\textwidth]{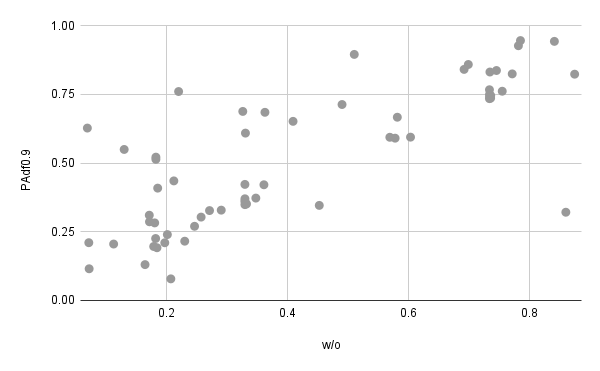}\label{fig:corr_wo-padf}} \\
\subfigure[{~${\rm PAdf}_{0.9}$ and PA}]{ 
\includegraphics[width=0.8\textwidth]{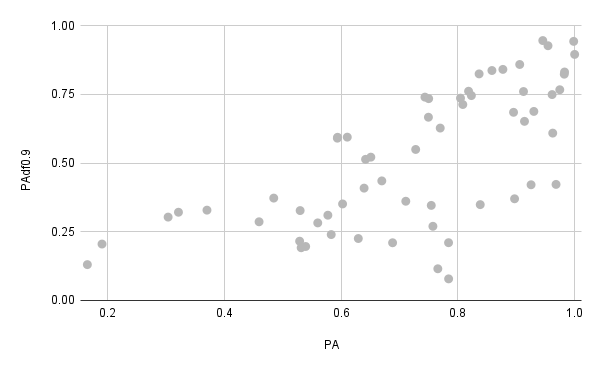}\label{fig:corr_pa-padf}} \\
\subfigure[{~${\rm PAdf}_{0.9}$ and PA\%K}]{
\includegraphics[width=0.8\textwidth]{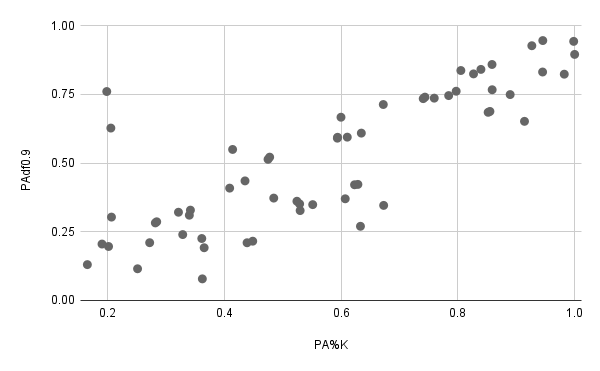}\label{fig:corr_pak-padf}} \\
  \end{center}
  \caption{This figure shows the correlations between the metrics : w/o, PA, PA\%K, PAdf.} \label{fig:corr}
\end{figure}

Additionally, we evaluate the performance of the 10 baselines on the 6 public datasets for the F1 scores without any protocols and with the PA, PA\%K, and PAdf protocols.
The experimental results are shown in Table~\ref{table:TSB-UAD1} and \ref{table:TSB-UAD2}.
It is worth noting that the datasets and models are included in TSB-UAD pipeline, so that
a detailed description of the datasets and baselines can be found in \cite{paparrizos2022tsb}.
The distribution charts of Table~\ref{table:TSB-UAD1} and \ref{table:TSB-UAD2} are shown in Fig.~\ref{fig:corr},
where the correlation coefficients are given as follows;
correlation between the F1-score without any protocols(w/o) and that of the PAdf: 0.77, correlation between the F1-score of the PA and that of the PAdf: 0.67, and correlation between the F1-score of the PA\%K and that of the PAdf: 0.83.
The overestimation problem of the PA protocol can be seen in Fig.~\ref{fig:corr_pa-padf}.
The F1 scores of ${\rm PAdf}_{0.9}$ are evenly distributed from 0 to 1, while the F1 scores of the PA are distributed in regions higher than 0.5.
Also, many points are located at high values on the PA axis, but at low values on the ${\rm PAdf}_{0.9}$ axis.

\begin{table*}[h]
\caption{The F1 scores without any protocols and with the PA/PA\%K/PAdf protocols measure for the 6 public datasets with respect to the 10 baselines. (Continue to Table~\ref{table:TSB-UAD2})}\label{table:TSB-UAD1}
\vskip 0.15in
\begin{center}
\begin{sc}
\resizebox{0.9\textwidth}{!}{%
\begin{tabular}{l||cccc}
\toprule
dataset & {} & Dodgers & {} & {}\\
\hline
\hline
 baseline & w/o & PA & PA\%K &  ${\rm PAdf}_{0.9}$\\
 \hline
SAND & 0.578 & 0.593 & 0.593 & 0.590 \\
POLY&0.165 & 0.165 & 0.165 & 0.130 \\
PCA & 0.603 & 0.610 & 0.610 & 0.594 \\
OCSVM & 0.291 & 0.370 & 0.341 & 0.328 \\
MatrixProfile & 0.569 & 0.594 & 0.594 & 0.594  \\
LSTM & 0.172 & 0.577 & 0.339 & 0.310 \\
LOF & 0.179 & 0.539 & 0.201 & 0.196 \\
IForest & 0.582 & 0.749 & 0.599 & 0.667 \\
CNN & 0.181 & 0.560 & 0.281 & 0.282 \\
AE & 0.172 & 0.459 & 0.284 & 0.286 \\
\midrule
dataset & {} & ECG & {}& {} \\
\hline
\hline
baseline & w/o & PA & PA\%K &  ${\rm PAdf}_{0.9}$\\
\hline
SAND & 0.841 & 0.998 & 0.998 & 0.942 \\
POLY & 0.860 & 0.321 & 0.321 & 0.321 \\
PCA & 0.782 & 0.954 & 0.927 & 0.927 \\
OCSVM & 0.699 & 0.906 & 0.858 & 0.858 \\
MatrixProfile & 0.201 & 0.583 & 0.328 & 0.240 \\
LSTM & 0.130 & 0.727 & 0.414 & 0.549 \\
LOF & 0.207 & 0.784 & 0.362 & 0.079 \\
IForest & 0.785 & 0.945 & 0.945 & 0.945 \\
CNN & 0.409 & 0.914 & 0.914 & 0.652 \\
AE & 0.692 & 0.877 & 0.839 & 0.840 \\
\midrule
dataset & {} & MITDB & {} & {} \\
\hline
\hline
baseline & w/o & PA & PA\%K & ${\rm PAdf}_{0.9}$\\
\hline
SAND & 0.875 & 0.982 & 0.982 & 0.823 \\
POLY & 0.113 & 0.190 & 0.190 & 0.205 \\
PCA & 0.220 & 0.912 & 0.198 & 0.760 \\
OCSVM & 0.363 & 0.895 & 0.852 & 0.684 \\
MatrixProfile & 0.510 & 1.000 & 1.000 & 0.895 \\
LSTM & 0.072 & 0.784 & 0.271 & 0.210 \\
LOF & 0.257 & 0.303 & 0.206 & 0.303 \\
IForest & 0.069 & 0.770 & 0.205 & 0.627 \\
CNN & 0.072 & 0.765 & 0.251 & 0.115 \\
AE & 0.326 & 0.930 & 0.855 & 0.688 \\
\bottomrule
\end{tabular} }
\end{sc}
\end{center}
\vskip -0.1in
\end{table*}

\begin{table*}[h]
\caption{Continuing from Table~\ref{table:TSB-UAD1}, This table shows that the F1 scores without any protocols and with the PA/PA\%K/PAdf protocols measure for the 6 public datasets with respect to the 10 baselines.}\label{table:TSB-UAD2}
\vskip 0.15in
\begin{center}
\begin{sc}
\resizebox{0.9\textwidth}{!}{%
\begin{tabular}{l||cccc}
\toprule
dataset & {} & SensorScope & {} & {} \\
\hline
\hline
 baseline & w/o & PA & PA\%K &  ${\rm PAdf}_{0.9}$ \\
 \hline
SAND & 0.332 & 0.602 & 0.529 & 0.351 \\
POLY & 0.348 & 0.484 & 0.484 & 0.372 \\
PCA & 0.453 & 0.754 & 0.673 & 0.346 \\
OCSVM & 0.361 & 0.925 & 0.623 & 0.421 \\
MatrixProfile & 0.330 & 0.838 & 0.551 & 0.349 \\
LSTM & 0.331 & 0.962 & 0.634 & 0.609 \\
LOF & 0.330 & 0.711 & 0.524 & 0.361 \\
IForest & 0.490 & 0.808 & 0.672 & 0.713 \\
CNN & 0.330 & 0.968 & 0.628 & 0.422 \\
AE & 0.330 & 0.897 & 0.607 & 0.370 \\
\midrule
dataset & {} & SVDB &  {}& {} \\
\hline
\hline
baseline & w/o & PA & PA\%K &  ${\rm PAdf}_{0.9}$ \\
\hline
SAND & 0.734 & 0.743 & 0.743 & 0.740 \\
POLY & 0.755 & 0.818 & 0.797 & 0.761 \\
PCA & 0.771 & 0.836 & 0.827 & 0.824 \\
OCSVM & 0.736 & 0.804 & 0.759 & 0.736 \\
MatrixProfile & 0.734 & 0.750 & 0.740 & 0.734 \\
LSTM & 0.734 & 0.961 & 0.889 & 0.749 \\
LOF & 0.736 & 0.823 & 0.784 & 0.745 \\
IForest  & 0.745 & 0.858 & 0.805 & 0.836 \\
CNN & 0.734 & 0.975 & 0.859 & 0.766 \\
AE  & 0.735 & 0.983 & 0.945 & 0.831 \\
\midrule
dataset &{} & NAB & {} & {} \\
\hline
\hline
baseline & w/o & PA & PA\%K & ${\rm PAdf}_{0.9}$ \\
\hline
SAND & 0.230 & 0.529 & 0.448 & 0.216 \\
POLY & 0.271 & 0.529 & 0.529 & 0.327 \\
PCA & 0.212 & 0.669 & 0.435 & 0.435 \\
OCSVM & 0.246 & 0.757 & 0.633 & 0.270 \\
MatrixProfile  & 0.184 & 0.531 & 0.365 & 0.192 \\
LSTM  & 0.183 & 0.651 & 0.477 & 0.521 \\
LOF  & 0.182 & 0.629 & 0.361 & 0.225 \\
IForest  & 0.186 & 0.639 & 0.409 & 0.409 \\
CNN  & 0.183 & 0.642 & 0.474 & 0.513 \\
AE  & 0.197 & 0.688 & 0.438 & 0.210 \\
\bottomrule
\end{tabular} }
\end{sc}
\end{center}
\vskip -0.1in
\end{table*}

\end{document}